\pdfoutput=1
\documentclass[11pt]{article}

\usepackage[]{acl}

\usepackage{times}
\usepackage{latexsym}

\usepackage[T1]{fontenc}
\usepackage[utf8]{inputenc}

\usepackage{microtype}

\usepackage{graphicx} %images
\usepackage{multirow} % for table
\usepackage{amsmath}%
\usepackage{MnSymbol}%
\usepackage{wasysym}%
\usepackage{tikz}

\usepackage{enumerate} % for table enumarete
\usepackage{paralist}
\usepackage{inconsolata}
\usepackage{booktabs}
\usepackage{mathtools}
\DeclareUnicodeCharacter{2061}{}
\title{It Is Not About What You Say, It Is About How You Say It: A Surprisingly Simple Approach for Improving Reading Comprehension}

\author{Sagi Shaier,$^\nabla$ Lawrence E Hunter,$^\dag$ Katharina von der Wense$^{\nabla\diamondsuit}$ \\
  $^\nabla$University of Colorado Boulder\\
$^\dag$Independent Scholar \\
$^\diamondsuit$Johannes Gutenberg University Mainz\\
$^\nabla$E-mail: \{sagi.shaier, katharina.kann\}@colorado.edu \\
$^\dag$E-mail: Prof.Larry.Hunter@gmail.com \\
 \\}

\begin{document}
\maketitle

\begin{abstract}
Natural language processing has seen rapid progress over the past decade. Due to the speed of developments, some practices get established without proper evaluation. Considering one such case and focusing on reading comprehension, we ask our first research question: 1) How does the order of inputs -- i.e., question and context -- affect model performance? Additionally, given recent advancements in input emphasis, we ask a second research question: 2) Does emphasizing either the question, the context, or both enhance performance? Experimenting with 9 large language models across 3 datasets, we find that presenting the context before the question improves model performance, with an accuracy increase of up to $31\%$. Furthermore, emphasizing the context yields superior results compared to question emphasis, and in general, emphasizing parts of the input is particularly effective for addressing questions that models lack the parametric knowledge to answer. Experimenting with both prompt-based and attention-based emphasis methods, we additionally find that the best method is surprisingly simple: it only requires concatenating a few tokens to the input and results in an accuracy improvement of up to $36\%$, allowing smaller models to outperform their significantly larger counterparts. 
\end{abstract}

\section{Introduction}
For the task of reading comprehension (RC), models receive two kinds of inputs: 1) a context, e.g., a Wikipedia article, and 2) a question that should be answered according to the context \cite{dzendzik-etal-2021-english, zeng2020survey}. While early efforts to address this task usually involve models that encode each of these separately \cite{zhang-2019-mc, tay-etal-2018-multi, nishida-etal-2019-answering, clark-gardner-2018-simple, choi-etal-2017-coarse}, more recently, large language models (LLMs) receive a concatenation of the two inputs \cite{wen-etal-2022-m3, huang-etal-2022-understand, sun-etal-2023-answering, shaier2024adaptivequestionansweringenhancing, bahak2023evaluating, baek-etal-2023-knowledge-augmented, NEURIPS2020_1457c0d6, chowdhery2022palm, chung2022scaling}.

\begin{figure}[t]
\centering
\includegraphics[width=1\columnwidth,height=0.7\columnwidth,keepaspectratio]{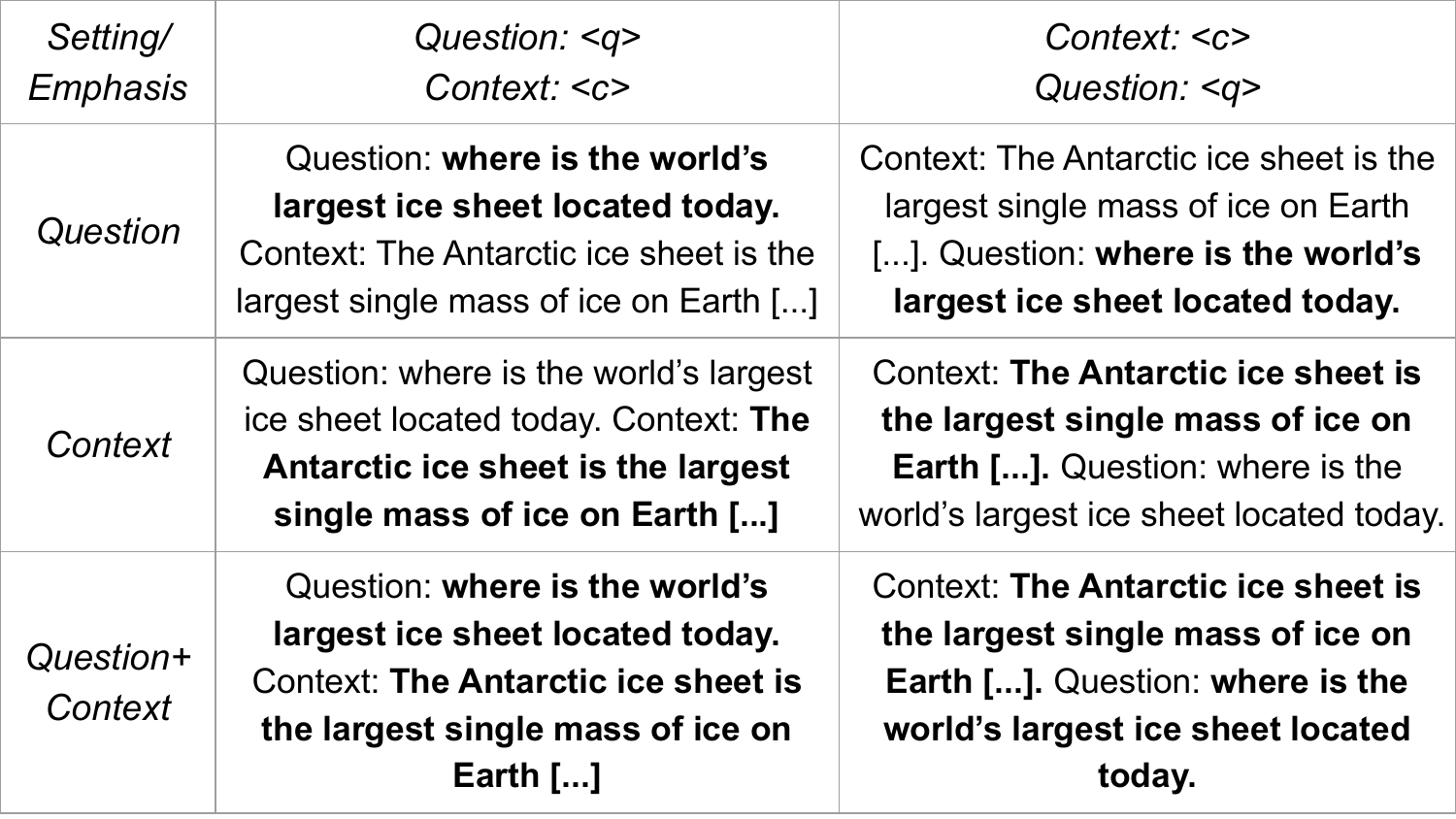}
\caption{Example from the Natural Questions dataset in which we show the different settings we experiment with: question or context first in the input prompt, and the different substring emphasis (in bold). <q>=question string; <c>=context string.}
\label{main_figure}
\end{figure}

% However, 
% while much of prompt development is based on intuition \cite{liu2021pretrain, gao-etal-2021-making, webson-pavlick-2022-prompt}, 
Surprisingly, \textbf{there is no current standard of what the ordering of such input components should be}. For example, \citet{sun-etal-2023-answering, nori2023capabilities, bahak2023evaluating, kamalloo-etal-2023-evaluating, singhal2022large, zhong-etal-2022-proqa} provide the question first in each prompt, while \citet{cheng2023adapting, nori2023capabilities, liu2023evaluating, baek-etal-2023-knowledge-augmented, NEURIPS2020_1457c0d6, singhal2022large, chowdhery2022palm, chung2022scaling} provide the context first. Moreover, \textbf{there is no current standard of how to present the two input components in general}. For example, considering the question and context strings <q> and <c>, respectively, \citet{wen-etal-2022-m3} add the special tokens “question:” and “context:” before the question and context, while \citet{nori2023capabilities} use “<c>**Question:** <q>”, \citet{zhong-etal-2022-proqa} use “[Question]: <q> [Passage]: <c>”, \citet{liu2023evaluating} use “<c> <q>”, and others such as \cite{baek-etal-2023-knowledge-augmented, NEURIPS2020_1457c0d6, chowdhery2022palm, chung2022scaling}, employ their own methods.

While at first sight this might not seem important, many works have shown that LMs can be extremely susceptible to slight variations in the input sequence \cite{jia-liang-2017-adversarial, si2019does, sen-saffari-2020-models, shaier-etal-2023-emerging}. Furthermore, recent research has found that \textbf{different presentations of inputs can help emphasize them} and 
% tokens in the input sequence, often by presenting them differently, can 
improve models' ability to follow instructions \cite{zhang2023tell}. Based on these observations, we ask the following research questions (RQs): 1) How does the order of inputs – i.e., question and context – affect model performance? 2) Does emphasizing either the question, the context, or both enhance performance? A summary of these questions can be seen in Figure \ref{main_figure}.

We evaluate 9 LLMs on 3 datasets and find the following: 1) The ordering of the question and context is crucial, and improves model performance with an accuracy increase of up to $31\%$. 2) Both prompt-based and attention-based emphasis methods are capable of strongly improving models' performance, where emphasizing the context yields superior results compared to emphasizing the question, and in general, emphasizing parts of the input is particularly effective for addressing questions that models lack the parametric knowledge to answer. 3) The best emphasis method is surprisingly simple: it only requires a simple concatenation of a few tokens to the input and results in an accuracy improvement of up to $36\%$, allowing smaller models to outperform their significantly larger counterparts.

\section{Related Work}
\label{related_work}
\paragraph{Reading Comprehension}
% In the RC task, models receive two kinds of inputs: 1) a context, and 2) a question that should be answered according to the context \cite{dzendzik-etal-2021-english, zeng2020survey, lai-etal-2017-race, kocisky-etal-2018-narrativeqa, rajpurkar-etal-2016-squad}. Early approaches to handle such inputs involved models that encode each of these separately \cite{zhang-2019-mc, tay-etal-2018-multi, nishida-etal-2019-answering, clark-gardner-2018-simple, choi-etal-2017-coarse, kadlec-etal-2016-text, yu2018qanet, seo2018bidirectional, xiong2018dynamic}. 

Reading comprehension involves the task of understanding a given context, such as a Wikipedia passage and answering questions based on that context \cite{dzendzik-etal-2021-english, zeng2020survey}. In comparison to answering questions based on models' parametric knowledge \cite{petroni2019languagemodelsknowledgebases, geva2021transformerfeedforwardlayerskeyvalue, shaier-etal-2024-comparing, roberts-etal-2020-much}. To that end, researchers develop models capable of comprehending written text and extracting relevant information to accurately respond to queries \cite{yang-etal-2019-enhancing-pre, wang-pan-2022-deep, touvron2023llama}. Traditional approaches often encode the context and question separately \cite{zhang-2019-mc, tay-etal-2018-multi, nishida-etal-2019-answering, clark-gardner-2018-simple, choi-etal-2017-coarse}, while more recent advancements leverage LLMs that concatenate both inputs into a single string \cite{wen-etal-2022-m3, huang-etal-2022-understand, sun-etal-2023-answering, bahak2023evaluating, baek-etal-2023-knowledge-augmented, NEURIPS2020_1457c0d6, chowdhery2022palm, chung2022scaling}. These models need to possess a deep understanding of the provided context to generate accurate responses to a wide range of questions, and many have shown that they do. Achieving high performance in reading comprehension tasks requires not only effective encoding of textual information, but also sophisticated reasoning and inference abilities to derive answers from the context accurately \cite{xie-xing-2017-constituent}. Therefore, ongoing research on reading comprehension focuses on improving model architectures \cite{dhingra-etal-2017-gated, indurthi-etal-2018-cut, wang-pan-2022-deep, touvron2023llama}, training strategies \cite{gottumukkala-etal-2020-dynamic, xu-etal-2019-multi}, and evaluation metrics \cite{yang-etal-2018-adaptations,sugawara-etal-2017-evaluation} to enhance the comprehension and reasoning capabilities of these systems. Here, we address the gap in research focused on how the inputs themselves can impact performance.

\paragraph{Prompt Engineering}
A related area -- prompt engineering \cite{strobelt2022interactive, bach-etal-2022-promptsource} -- focuses on modifying the input prompt to improve the performance of LMs without altering their underlying architecture or training regime. 
% And while LMs require a deep understanding of the provided context to generate accurate responses, 
Recent studies have demonstrated that large performance enhancements can be achieved through prompt engineering alone \cite{NEURIPS2020_1457c0d6, liu2021pretrain, wei2023chainofthought, dong2023survey}. This approach involves various techniques such as adding different input strings \cite{zhang2023tell}, providing step-by-step instructions \cite{wei2023chainofthought}, or incorporating additional contextual information into the prompt \cite{NEURIPS2020_1457c0d6}. By carefully crafting the input prompt, researchers aim to guide the model towards relevant information and improve its ability to comprehend and generate coherent responses. 
% Here, we modify the presentation of the question and context in the prompt by either: 1) changing the question and context location, or 2) emphasizing them, e.g., by adding a string before and after the question.

% Prompt engineering offers a straightforward yet effective method for fine-tuning LMs for specific tasks or domains without the need for extensive architectural modifications or retraining procedures. 

\paragraph{Emphasis Methods}
It is important to note that researchers often do not have the ability to precisely guide the model using prompt engineering, and much of prompt development is based on intuition. That is, researchers often have to try many different prompts manually or automatically until they find those that increase performance, and often just for their specific models \cite{liu2021pretrain, gao-etal-2021-making, webson-pavlick-2022-prompt}. 
% Moreover, the reason for why such a method works in the first place is still an open question. 
In comparison, recent work on input emphasis, including attention steering (AS) and marked prompting (MP) \cite{zhang2023tell}, have shown great success in improving models' ability to follow instructions. These methods aim to guide the focus of models towards various segments of the input sequence, by either adding tokens to the sequence or rescaling attention weights for relevant tokens. AS is closely related to work that avoids modifying models' architectures, or training regime, however, it takes a more direct approach by modifying parts of the input directly by rescaling the attention values of specific heads corresponding to specific tokens. 

\paragraph{Interpretability}
Emphasis methods, such as AS, are also related to model interpretability, which is concerned with understanding the contributions of different model components, and in particular, head attribution \cite{geva-etal-2023-dissecting}. For example, \citet{meng2023locating, geva-etal-2021-transformer, kobayashi2023analyzing} show that different knowledge from the training data is found within the feedforward layers, \citet{shaier-etal-2023-stochastic} discuss the role of the
training data in explaining model generation, while others show that attention heads have similar patterns \cite{geva-etal-2023-dissecting}. 

\section{Models}
We experiment with 9 different LLMs. 

\paragraph{Llama-2-7B and Llama-2-13B} Llama-2-7B and Llama-2-13B \cite{touvron2023llama} are LLMs which contain 7 and 13 billion parameters, respectively, and are trained on 2 trillion tokens. We use these models as they perform well on the reading comprehension task \cite{touvron2023llama} and recent work shows that their performance can be improved using emphasis methods \cite{zhang2023tell}.

\paragraph{Falcon-7B and Falcon-7B Instruct} These two models contain 7 billion parameters each, and are trained on 1.5 trillion tokens \cite{almazrouei2023falcon}. We opt for these models because they are newer and have demonstrated significant success across various tasks. Additionally, Falcon-7B Instruct comes with an instruct version, enabling us to compare the performance of both variations.

\paragraph{MPT-7B and MPT-7B Instruct} These are two LLMs with 7 billion parameters, trained on 1 trillion tokens \cite{mpt_2023}. Chosen for their recent development and proven versatility.
% , MPT-7B Instruct offers an instruct version, facilitating performance comparison between the two variants.

\paragraph{GPT-J-6B} GPT-J-6B \cite{gpt-j} 
% is a model from the GPT model family. It 
contains 6 billion parameters and is trained on the Pile dataset \cite{gao2020pile}. We use this model as in addition to the fact that it has been shown to perform well on question answering tasks \cite{de-bruyn-etal-2022-smaller}, it is also often compared again our largest model -- Llama-2 \cite{touvron2023llama, zhang2023tell} and recent work shows that its performance can be improved using emphasis methods \cite{zhang2023tell}.

\paragraph{GPT-2-XL} GPT-2-XL \cite{Radford2019LanguageMA} a LLM with 1.5 billion parameters and is trained on WebText \cite{Radford2019LanguageMA}. While much smaller than current state-of-the-art models, such as ChatGPT \cite{openai_2023} or GPT 4 \cite{openai2023gpt4}, we experiment with it as many low-resource settings require usage of smaller models.
% we experiment with it to explore whether our research questions also generalize to smaller model variants.

\paragraph{GPT-2-Large} Our last model, GPT-2-Large \cite{Radford2019LanguageMA}, contains 774 million parameters and, similar to GPT-2-XL, is trained on WebText. We use it for similar reasons as those we described in the GPT-2-XL Section.

\section{Experiments}

\subsection{Datasets}
We experiment with the following RC datasets: 

\paragraph{Natural Questions} The natural questions dataset \cite{kwiatkowski-etal-2019-natural} is comprised of authentic, anonymized, and aggregated queries directed to the Google search engine. Each question is accompanied by an entire Wikipedia page, and a collection of annotated long and short answers. As entire Wikipedia pages exceed many of our models' context lengths, for each question, we use each of the long answers as the context and the corresponding short answers as the gold answers.

We utilize it due to its widespread adoption and popularity within the research community, ensuring the reproducibility and comparability of our results with existing studies. Additionally, its comprehensive coverage of diverse question types and real-world contexts allows us to further evaluate whether our findings generalize. 

\paragraph{Stanford Question Answering Dataset (SQuAD)} SQuAD \cite{rajpurkar-etal-2016-squad} is composed of questions that are gathered from crowdworkers who ask questions about Wikipedia articles. We choose to use it for similar reasons described as the Natural Questions dataset.\footnote{We use the 1.0 version instead of the 2.0 version, as the later version contains empty strings as labels for its irrelevant contexts, which prevents us from using the closed-book setting to determine its parametric knowledge (see Section \ref{knowledge_amount}).}

\paragraph{AdversarialQA} The AdversarialQA dataset \cite{bartolo-etal-2020-beat} has been constructed adversarially, based on 3 models-in-the-loop. More specifically, the authors use the same SQuAD annotation methodology and models trained on it, and explore an annotation setting where annotators are tasked with formulating questions for which the model yields incorrect predictions. Consequently, the dataset is composed solely of instances where models answer inaccurately. While not as popular as SQuAD or the Natural Questions, we utilize this dataset as this annotation methodology makes these questions unique and especially challenging.

% and 4) MedHop \cite{welbl-etal-2018-constructing}, a multihop biomedical QA dataset that has questions on drug--drug interactions from MEDLINE abstracts. We focus our main experiments on the first 3 datasets (Natural Questions, SQuAD, and AdversarialQA), and use the last one (MedHop) to show that attention steering also generalizes to unseen datasets in different domains, such as biomedicine. 

\paragraph{Data Splits} As the test set for each of these datasets is either private or does not contain gold answers, we randomly split the validation sets into two parts and use one half as our validation set and the other as our held-out test set. This results in roughly the following split for each dataset. Natural Questions: $307k$ train, $3915$ validation, $3915$ test, SQuAD: $87k$ train, $5285$ validation, $5285$ test, AdversarialQA: $30k$ train, $1500$ validation, $1500$ test.

% \subsection{Setting}
\subsection{Prompt Structure}
\label{setting}
RC datasets consist of question, context, and answer triples $(q,c, a)$, where $q \in Q$, $c \in C$, $a \in A$.
% two main components: 1) a collection of contexts, $C$, and 2) questions about the contexts, $Q$. 
As outlined above, our RQ1 is concerned with the order in which the question and context are provided to the model: since previous work has been inconsistent in this regard, we explore which order (if any) results in higher performance. 

Concretely, we compare the following two prompt structures (cf. Figure \ref{main_figure}): 

\paragraph{Question First} Here, the question comes first in the prompt. In our concrete format, this results in the input sequence 
\begin{center}
\texttt{
Question: <q> Context: <c>,   
}
\end{center}
where $q$ and $c$ are pairs of question and context strings, $q \in Q$, $c \in C$.

\paragraph{Context First} In this setting, the context is the first part of the prompt. In our concrete format, this results in the input sequence 
\begin{center}
\texttt{
Context: <c> Question: <q>,
}
\end{center}
where, again, $q$ and $c$ are question--context pairs, $q \in Q$, $c \in C$. 

% \subsection{Baselines}
\subsection{Emphasis Strategies}

\paragraph{Marked Prompting}
MP \cite{zhang2023tell} is a simple prompt-based approach in which we append a string to the input sequence in order to emphasize it. For example, to emphasize the questions, we can append the string \textbf{“ * ”} to
\begin{center}
Question: <q> Context: <c>    
\end{center}
which would result in 
\begin{center}
Question: \textbf{*}<q>\textbf{*} Context: <c>    
\end{center}

We experiment with 4 MP methods, composed of the following start and end string pairs: [* and *, “ and ”, <emphasize> and <$\backslash$emphasize>, <mark> and <$\backslash$mark>]

\paragraph{Attention Steering}
In comparison to MP, AS is a more computationally-intensive method to emphasize input tokens and is attention-based.

We follow \citet{zhang2023tell}'s approach known as PASTA, which requires 1) an LLM with $L$ stacked layers, each with $N$ multi-head attention (MHA) submodules, such as most transformer-based models \cite{NIPS2017_3f5ee243}; 2) input text $W$, and 3) a segment $w \in W$ that is found within the input text.

PASTA is composed of two parts: 

1) \textit{Attention steering}: in this part, we downweight the attention scores of any token that is not part of the segment $w$, by multiplying them with a small scalar $ 0 \leq \alpha < 1 $\, for a selected $n \in N$ MHA submodules. In our experiments, we use $\alpha=1e^{-3}$ based on \citet{zhang2023tell}.

2) \textit{Model profiling}: here, we select which $n \in N$ to apply the AS to. While the original paper experiments with several selection methods, such as applying the steering to all heads, single heads, or entire layers, they obtain the best performance when selecting the intersection of the top-k best performing heads across several datasets. 
% For example, if the top-5 heads for datasets 1 and 2 are $\{1,10,157, 563, 600\}$ and $\{157, 158, 600, 605, 700 \}$, respectively, the heads that will be steered are the intersection of these:  $n=\{157,600\}$. 
They select $k$ from a small number of options, such as $\{300, 400, 500\}$ for Llama 7B. However, we find that we can improve performance by increasing this range.

In particular, from each dataset's \textit{training split} $D_{ti}$, we take a small subset of examples $d_{ti} \in D_{ti}$, and apply AS to each head individually. In our experiments, we use $|d_{ti}|=1000$ for GPT-2 large and XL, and $|d_{ti}|=500$ for GPT-J and Llama-2, for computational reasons, after manually assessing different values which result in roughly similar models' scores. We store the performance of the model for each head, which results in $L * N$ scores for each $d_{ti}$. Next, on each dataset's \textit{validation split} $D_{vi}$ we iteratively select a $k$, where $0 < k \leq N*L$, and find the intersection of the top-k performing heads across all datasets $d_{ti} \in D_{ti}$.
We store the scores, which results in $L * N$ scores for each $k$ for each $D_{vi}$. For the test split, we use the best $k$ based on the validation split.

\paragraph{Baseline: No Emphasis}
As a baseline, we further compare to a setting in which we do not emphasize any string and use the original prompt from Section \ref{setting} as inputs to the models. 

\subsection{Hyperparameters}
%As we evaluate different models which have different tokenizers, some sequences are longer than some models' maximum sequence length. This is problematic as we experiment with 1) different models; 2) different prompt structures. The former may result in unfair comparison between models, 
We use a maximum sequence length of 512. Truncation due to this might result in an unfair comparison between the different prompt structures as either question or context might get truncated.\footnote{See Section \ref{additional_models_sec} for an analysis of models with a larger context length.}

%if we place the question last, and some models will truncate some of the strings, we would not be able to infer if the different performance is a result of the truncation or the actual prompt structure. Hence, to ensure a fair comparison between the different models' tokenizers, 
% do no truncate part of the sequences differently, which would skew the results for the two settings, 
In order to avoid this, we remove sequences that are longer than 512 tokens
%, which all models can handle and do not remove a lot of the data 
(about $15\%$ of the examples in the Natural Questions dataset, less than $1\%$ for SQuAD, and $0\%$ for AdversarialQA). 

% However, we also experiment with 5 additional LLMs on the Natural Questions dataset, all of which were published in 2023 or afterwards and contain between 7B and 13B parameters. 2 of the 5 additional LLMs were instruction-tuned, and all 5 of the additional models were evaluated using their maximum context size (up to 4k). Our results can be seen in Appendix \ref{max_context_size}.

% \subsection{Evaluation Metrics}
\subsection{Metrics}
\paragraph{Accuracy}
Following \citet{liu2023lost, kandpal2023large, mallen-etal-2023-trust}, we assess the performance of all models using accuracy, determining if any of the gold responses are present in the predicted output. Concretely, we feed the two prompts described in Section \ref{setting}, such as \textit{“Question: <q>. Context: <c>”}, to each of the models, and evaluate whether the gold label answer exists within the LLM generated answer.\footnote{While this approach is popular, it is important to note that no existing evaluation metric is flawless. For instance, this approach may overlook accurate responses (e.g., because they are not an exact match to gold answers) or erroneously categorize incorrect responses as correct.
% no current evaluation metric is perfect. For example, this approach might miss correct answers or count answers as correct that are not. 
To address this concern, we supplement our evaluation process by manually inspecting 100 responses from Llama 2 on the Natural Questions dataset in the no emphasis, context-first setting, to evaluate the frequency of such occurrences. We find that while this approach identifies $58.1\%$ of the answers as correct, manual analysis identifies $82\%$. This highlights the gap between this popular method and human evaluation.
% may not be the best method to analyze model responses.
% To that end, we also add an analysis of how often that happens by manually inspecting 100 answers (see Appendix XXX).
}
\paragraph{Context-free Accuracy}
\label{knowledge_amount}
We are further interested in evaluating the models' parametric knowledge. For this, we follow work by \citet{shaier-etal-2024-desiderata, control, xie2023adaptive, roberts-etal-2020-much}, who use a closed-book setting to evaluate models' parametric knowledge. In particular, we define \textit{known knowledge} as questions that models answers correctly without the corresponding context and \textit{unknown knowledge} as those they cannot. 

\paragraph{Perplexity}
Perplexity (PPL) is defined as the exponentiated average of the negative log-likelihood of a sequence. Concretely, given a sequence of tokens $X=(x_0,x_1,...,x_t)$, the perplexity of $X$ denoted as 

\begin{center}
$PPL(X)=exp⁡( -\frac{1}{t} \sum_{i}^{t}
log⁡p_{\theta}(x_i \mid x_{<i}))$
\end{center}

where $log⁡p_{\theta}(x_i\mid x_{<i})$ represents the log-likelihood of the i-th token conditioned on the preceding tokens $x_{<i}$ according to the model. 

\section{Results}

\subsection{RQ 1: Question First vs. Context First}
\label{rq1_results}

%%%%%
\begin{table*}[h]
\centering
\tiny
\setlength{\tabcolsep}{1.0pt}
\begin{tabular}{|c|cc|cccccccc|c|cccccccc|c|cccccccc|}
\cline{1-11} \cline{13-20} \cline{22-29}
\textbf{Model}                                                         & \multicolumn{2}{c|}{\textbf{Emphasis Method}}                                        & \multicolumn{8}{c|}{\textbf{Natural Questions}}                                                                                                                                                                                                                                                                                                                                                                                                                   &  & \multicolumn{8}{c|}{\textbf{SQuAD}}                                                                                                                                                                                                                                                                                                                                                                                                                      &  & \multicolumn{8}{c|}{\textbf{AdversarialQA}}                                                                                                                                                                                                                                                                                                                                                                                                              \\ \cline{1-11} \cline{13-20} \cline{22-29} 
                                                                       &                                          &                                    & \multicolumn{4}{c|}{Question First}                                                                                                                                                    & \multicolumn{4}{c|}{Context First}                                                                                                                      &  & \multicolumn{4}{c|}{Question First}                                                                                                                                           & \multicolumn{4}{c|}{Context First}                                                                                                                      &  & \multicolumn{4}{c|}{Question First}                                                                                                                                                    & \multicolumn{4}{c|}{Context First}                                                                                                             \\ \cline{4-11} \cline{13-20} \cline{22-29} 
                                                                       &                                          &                                    & \multicolumn{1}{c|}{\multirow{2}{*}{\begin{tabular}[c]{@{}c@{}}No\\ Emphasis\end{tabular}}} & \multicolumn{3}{c|}{Emphasis}                                                                                                                    & \multicolumn{1}{c|}{\multirow{2}{*}{\begin{tabular}[c]{@{}c@{}}No\\ Emphasis\end{tabular}}} & \multicolumn{3}{c|}{Emphasis}                                                                                      &  & \multicolumn{1}{c|}{\multirow{2}{*}{\begin{tabular}[c]{@{}c@{}}No\\ Emphasis\end{tabular}}} & \multicolumn{3}{c|}{Emphasis}                                                                                                           & \multicolumn{1}{c|}{\multirow{2}{*}{\begin{tabular}[c]{@{}c@{}}No\\ Emphasis\end{tabular}}} & \multicolumn{3}{c|}{Emphasis}                                                                                      &  & \multicolumn{1}{c|}{\multirow{2}{*}{\begin{tabular}[c]{@{}c@{}}No\\ Emphasis\end{tabular}}} & \multicolumn{3}{c|}{Emphasis}                                                                                                                    & \multicolumn{1}{c|}{\multirow{2}{*}{\begin{tabular}[c]{@{}c@{}}No\\ Emphasis\end{tabular}}} & \multicolumn{3}{c|}{Emphasis}                                                                             \\ \cline{5-7} \cline{9-11} \cline{14-16} \cline{18-20} \cline{23-25} \cline{27-29} 
                                                                       &                                          &                                    & \multicolumn{1}{c|}{}                                                                       & \multicolumn{1}{c|}{Q}             & \multicolumn{1}{c|}{C}             & \multicolumn{1}{c|}{\begin{tabular}[c]{@{}c@{}}Q\\ +\\ C\end{tabular}} & \multicolumn{1}{c|}{}                                                                       & \multicolumn{1}{c|}{Q}             & \multicolumn{1}{c|}{C}    & \begin{tabular}[c]{@{}c@{}}Q\\ +\\ C\end{tabular} &  & \multicolumn{1}{c|}{}                                                                       & \multicolumn{1}{c|}{Q}    & \multicolumn{1}{c|}{C}             & \multicolumn{1}{c|}{\begin{tabular}[c]{@{}c@{}}Q\\ +\\ C\end{tabular}} & \multicolumn{1}{c|}{}                                                                       & \multicolumn{1}{c|}{Q}    & \multicolumn{1}{c|}{C}             & \begin{tabular}[c]{@{}c@{}}Q\\ +\\ C\end{tabular} &  & \multicolumn{1}{c|}{}                                                                       & \multicolumn{1}{c|}{Q}             & \multicolumn{1}{c|}{C}             & \multicolumn{1}{c|}{\begin{tabular}[c]{@{}c@{}}Q\\ +\\ C\end{tabular}} & \multicolumn{1}{c|}{}                                                                       & \multicolumn{1}{c|}{Q}    & \multicolumn{1}{c|}{C}    & \begin{tabular}[c]{@{}c@{}}Q\\ +\\ C\end{tabular} \\ \cline{1-11} \cline{13-20} \cline{22-29} 
\multirow{6}{*}{Llama-2}                                               & \multicolumn{1}{c|}{B}                  &                                    & \multicolumn{1}{c|}{46.3}                                                                   & \multicolumn{1}{c|}{}              & \multicolumn{1}{c|}{}              & \multicolumn{1}{c|}{}                                                  & \multicolumn{1}{c|}{58.1}                                                                   & \multicolumn{1}{c|}{}              & \multicolumn{1}{c|}{}     &                                                   &  & \multicolumn{1}{c|}{60.4}                                                                   & \multicolumn{1}{c|}{}     & \multicolumn{1}{c|}{}              & \multicolumn{1}{c|}{}                                                  & \multicolumn{1}{c|}{72.9}                                                                   & \multicolumn{1}{c|}{}     & \multicolumn{1}{c|}{}              &                                                   &  & \multicolumn{1}{c|}{42.6}                                                                   & \multicolumn{1}{c|}{}              & \multicolumn{1}{c|}{}              & \multicolumn{1}{c|}{}                                                  & \multicolumn{1}{c|}{49.4}                                                                   & \multicolumn{1}{c|}{}     & \multicolumn{1}{c|}{}     &                                                   \\ \cline{2-11} \cline{13-20} \cline{22-29} 
                                                                       & \multicolumn{1}{c|}{AS}                  &                                    & \multicolumn{1}{c|}{}                                                                       & \multicolumn{1}{c|}{54.8}          & \multicolumn{1}{c|}{53.0}          & \multicolumn{1}{c|}{-}                                                 & \multicolumn{1}{c|}{}                                                                       & \multicolumn{1}{c|}{57.8}          & \multicolumn{1}{c|}{59.3} & -                                                 &  & \multicolumn{1}{c|}{}                                                                       & \multicolumn{1}{c|}{66.3} & \multicolumn{1}{c|}{62.0}          & \multicolumn{1}{c|}{-}                                                 & \multicolumn{1}{c|}{}                                                                       & \multicolumn{1}{c|}{74.5} & \multicolumn{1}{c|}{72.9}          & -                                                 &  & \multicolumn{1}{c|}{}                                                                       & \multicolumn{1}{c|}{43.3}          & \multicolumn{1}{c|}{43.0}          & \multicolumn{1}{c|}{-}                                                 & \multicolumn{1}{c|}{}                                                                       & \multicolumn{1}{c|}{54.4} & \multicolumn{1}{c|}{53.3} & -                                                 \\ \cline{2-11} \cline{13-20} \cline{22-29} 
                                                                       & \multicolumn{1}{c|}{\multirow{4}{*}{MP}} & $\star$                            & \multicolumn{1}{c|}{}                                                                       & \multicolumn{1}{c|}{51.4}          & \multicolumn{1}{c|}{31.6}          & \multicolumn{1}{c|}{53.1}                                              & \multicolumn{1}{c|}{}                                                                       & \multicolumn{1}{c|}{58.3}          & \multicolumn{1}{c|}{56.4} & 58.8                                              &  & \multicolumn{1}{c|}{}                                                                       & \multicolumn{1}{c|}{56.8} & \multicolumn{1}{c|}{61.7}          & \multicolumn{1}{c|}{67.9}                                              & \multicolumn{1}{c|}{}                                                                       & \multicolumn{1}{c|}{69.1} & \multicolumn{1}{c|}{76.4}          & 79.7                                              &  & \multicolumn{1}{c|}{}                                                                       & \multicolumn{1}{c|}{40.5}          & \multicolumn{1}{c|}{43.2}          & \multicolumn{1}{c|}{46.7}                                              & \multicolumn{1}{c|}{}                                                                       & \multicolumn{1}{c|}{51.1} & \multicolumn{1}{c|}{54.2} & 57.5                                              \\ \cline{3-11} \cline{13-20} \cline{22-29} 
                                                                       & \multicolumn{1}{c|}{}                    & "                                  & \multicolumn{1}{c|}{}                                                                       & \multicolumn{1}{c|}{48.7}          & \multicolumn{1}{c|}{54.2}          & \multicolumn{1}{c|}{54.2}                                              & \multicolumn{1}{c|}{}                                                                       & \multicolumn{1}{c|}{56.4}          & \multicolumn{1}{c|}{58.2} & 59.9                                              &  & \multicolumn{1}{c|}{}                                                                       & \multicolumn{1}{c|}{61.4} & \multicolumn{1}{c|}{71.9}          & \multicolumn{1}{c|}{\underline{72.3}}                                              & \multicolumn{1}{c|}{}                                                                       & \multicolumn{1}{c|}{72.5} & \multicolumn{1}{c|}{76.3}          & 78.6                                              &  & \multicolumn{1}{c|}{}                                                                       & \multicolumn{1}{c|}{42.0}          & \multicolumn{1}{c|}{48.3}          & \multicolumn{1}{c|}{48.9}                                              & \multicolumn{1}{c|}{}                                                                       & \multicolumn{1}{c|}{50.2} & \multicolumn{1}{c|}{56.8} & 56.0                                              \\ \cline{3-11} \cline{13-20} \cline{22-29} 
                                                                       & \multicolumn{1}{c|}{}                    & <mark>      & \multicolumn{1}{c|}{}                                                                       & \multicolumn{1}{c|}{51.7}          & \multicolumn{1}{c|}{54.1}          & \multicolumn{1}{c|}{\underline{55.1}}                                              & \multicolumn{1}{c|}{\textbf{}}                                                              & \multicolumn{1}{c|}{60.0}          & \multicolumn{1}{c|}{55.5} & 60.5                                              &  & \multicolumn{1}{c|}{}                                                                       & \multicolumn{1}{c|}{53.3} & \multicolumn{1}{c|}{71.5}          & \multicolumn{1}{c|}{71.8}                                              & \multicolumn{1}{c|}{}                                                                       & \multicolumn{1}{c|}{75.4} & \multicolumn{1}{c|}{71.3}          & 80.4                                              &  & \multicolumn{1}{c|}{}                                                                       & \multicolumn{1}{c|}{39.0}          & \multicolumn{1}{c|}{47.3}          & \multicolumn{1}{c|}{\underline{49.3}}                                              & \multicolumn{1}{c|}{}                                                                       & \multicolumn{1}{c|}{50.7} & \multicolumn{1}{c|}{52.4} & \textbf{57.7}                                     \\ \cline{3-11} \cline{13-20} \cline{22-29} 
                                                                       & \multicolumn{1}{c|}{}                    & <emphasize> & \multicolumn{1}{c|}{}                                                                       & \multicolumn{1}{c|}{47.6}          & \multicolumn{1}{c|}{54.4}          & \multicolumn{1}{c|}{53.9}                                              & \multicolumn{1}{c|}{}                                                                       & \multicolumn{1}{c|}{\textbf{61.3}} & \multicolumn{1}{c|}{55.5} & 60.2                                              &  & \multicolumn{1}{c|}{}                                                                       & \multicolumn{1}{c|}{53.8} & \multicolumn{1}{c|}{72.2}          & \multicolumn{1}{c|}{68.0}                                              & \multicolumn{1}{c|}{}                                                                       & \multicolumn{1}{c|}{78.1} & \multicolumn{1}{c|}{70.4}          & \textbf{81.5}                                     &  & \multicolumn{1}{c|}{}                                                                       & \multicolumn{1}{c|}{37.8}          & \multicolumn{1}{c|}{\underline{49.3}}          & \multicolumn{1}{c|}{46.5}                                              & \multicolumn{1}{c|}{}                                                                       & \multicolumn{1}{c|}{51.4} & \multicolumn{1}{c|}{50.4} & 56.2                                              \\ \cline{1-11} \cline{13-20} \cline{22-29} 
\multirow{6}{*}{GPT-J}                                                 & \multicolumn{1}{c|}{B}                  &                                    & \multicolumn{1}{c|}{33.3}                                                                   & \multicolumn{1}{c|}{}              & \multicolumn{1}{c|}{}              & \multicolumn{1}{c|}{}                                                  & \multicolumn{1}{c|}{64.5}                                                                   & \multicolumn{1}{c|}{}              & \multicolumn{1}{c|}{}     &                                                   &  & \multicolumn{1}{c|}{45.5}                                                                   & \multicolumn{1}{c|}{}     & \multicolumn{1}{c|}{}              & \multicolumn{1}{c|}{}                                                  & \multicolumn{1}{c|}{61.0}                                                                   & \multicolumn{1}{c|}{}     & \multicolumn{1}{c|}{}              &                                                   &  & \multicolumn{1}{c|}{47.2}                                                                   & \multicolumn{1}{c|}{}              & \multicolumn{1}{c|}{}              & \multicolumn{1}{c|}{}                                                  & \multicolumn{1}{c|}{46.2}                                                                   & \multicolumn{1}{c|}{}     & \multicolumn{1}{c|}{}     &                                                   \\ \cline{2-11} \cline{13-20} \cline{22-29} 
                                                                       & \multicolumn{1}{c|}{AS}                  &                                    & \multicolumn{1}{c|}{}                                                                       & \multicolumn{1}{c|}{66.3}          & \multicolumn{1}{c|}{66.3}          & \multicolumn{1}{c|}{-}                                                 & \multicolumn{1}{c|}{}                                                                       & \multicolumn{1}{c|}{61.1}          & \multicolumn{1}{c|}{53.0} & -                                                 &  & \multicolumn{1}{c|}{}                                                                       & \multicolumn{1}{c|}{51.0} & \multicolumn{1}{c|}{44.6}          & \multicolumn{1}{c|}{-}                                                 & \multicolumn{1}{c|}{}                                                                       & \multicolumn{1}{c|}{55.8} & \multicolumn{1}{c|}{54.1}          & -                                                 &  & \multicolumn{1}{c|}{}                                                                       & \multicolumn{1}{c|}{45.0}          & \multicolumn{1}{c|}{37.8}          & \multicolumn{1}{c|}{-}                                                 & \multicolumn{1}{c|}{}                                                                       & \multicolumn{1}{c|}{41.6} & \multicolumn{1}{c|}{41.7} & -                                                 \\ \cline{2-11} \cline{13-20} \cline{22-29} 
                                                                       & \multicolumn{1}{c|}{\multirow{4}{*}{MP}} & $\star$                            & \multicolumn{1}{c|}{}                                                                       & \multicolumn{1}{c|}{33.4}          & \multicolumn{1}{c|}{26.9}          & \multicolumn{1}{c|}{49.7}                                              & \multicolumn{1}{c|}{}                                                                       & \multicolumn{1}{c|}{60.5}          & \multicolumn{1}{c|}{65.1} & 64.9                                              &  & \multicolumn{1}{c|}{}                                                                       & \multicolumn{1}{c|}{38.0} & \multicolumn{1}{c|}{52.5}          & \multicolumn{1}{c|}{41.7}                                              & \multicolumn{1}{c|}{}                                                                       & \multicolumn{1}{c|}{51.1} & \multicolumn{1}{c|}{64.0}          & 50.5                                              &  & \multicolumn{1}{c|}{}                                                                       & \multicolumn{1}{c|}{38.2}          & \multicolumn{1}{c|}{52.0}          & \multicolumn{1}{c|}{40.8}                                              & \multicolumn{1}{c|}{}                                                                       & \multicolumn{1}{c|}{40.2} & \multicolumn{1}{c|}{50.0} & 38.2                                              \\ \cline{3-11} \cline{13-20} \cline{22-29} 
                                                                       & \multicolumn{1}{c|}{}                    & "                                  & \multicolumn{1}{c|}{}                                                                       & \multicolumn{1}{c|}{39.0}          & \multicolumn{1}{c|}{63.0}          & \multicolumn{1}{c|}{62.3}                                              & \multicolumn{1}{c|}{}                                                                       & \multicolumn{1}{c|}{66.3}          & \multicolumn{1}{c|}{65.9} & 66.7                                              &  & \multicolumn{1}{c|}{}                                                                       & \multicolumn{1}{c|}{34.0} & \multicolumn{1}{c|}{56.2}          & \multicolumn{1}{c|}{49.5}                                              & \multicolumn{1}{c|}{}                                                                       & \multicolumn{1}{c|}{61.7} & \multicolumn{1}{c|}{61.0}          & 66.4                                              &  & \multicolumn{1}{c|}{}                                                                       & \multicolumn{1}{c|}{35.8}          & \multicolumn{1}{c|}{53.4}          & \multicolumn{1}{c|}{50.2}                                              & \multicolumn{1}{c|}{}                                                                       & \multicolumn{1}{c|}{48.7} & \multicolumn{1}{c|}{49.7} & 52.5                                              \\ \cline{3-11} \cline{13-20} \cline{22-29} 
                                                                       & \multicolumn{1}{c|}{}                    & <mark>      & \multicolumn{1}{c|}{}                                                                       & \multicolumn{1}{c|}{34.3}          & \multicolumn{1}{c|}{61.6}          & \multicolumn{1}{c|}{52.9}                                              & \multicolumn{1}{c|}{}                                                                       & \multicolumn{1}{c|}{61.5}          & \multicolumn{1}{c|}{\underline{67.8}} & 64.4                                              &  & \multicolumn{1}{c|}{}                                                                       & \multicolumn{1}{c|}{40.5} & \multicolumn{1}{c|}{64.2}          & \multicolumn{1}{c|}{55.9}                                              & \multicolumn{1}{c|}{}                                                                       & \multicolumn{1}{c|}{66.8} & \multicolumn{1}{c|}{68.5}          & \textbf{72.3}                                     &  & \multicolumn{1}{c|}{}                                                                       & \multicolumn{1}{c|}{41.8}          & \multicolumn{1}{c|}{64.1}          & \multicolumn{1}{c|}{52.2}                                              & \multicolumn{1}{c|}{}                                                                       & \multicolumn{1}{c|}{57.4} & \multicolumn{1}{c|}{55.0} & \underline{60.2}                                              \\ \cline{3-11} \cline{13-20} \cline{22-29} 
                                                                       & \multicolumn{1}{c|}{}                    & <emphasize> & \multicolumn{1}{c|}{}                                                                       & \multicolumn{1}{c|}{38.3}          & \multicolumn{1}{c|}{\textbf{69.0}} & \multicolumn{1}{c|}{64.2}                                              & \multicolumn{1}{c|}{}                                                                       & \multicolumn{1}{c|}{62.7}          & \multicolumn{1}{c|}{63.6} & 62.9                                              &  & \multicolumn{1}{c|}{}                                                                       & \multicolumn{1}{c|}{37.1} & \multicolumn{1}{c|}{\underline{64.7}}          & \multicolumn{1}{c|}{55.9}                                              & \multicolumn{1}{c|}{}                                                                       & \multicolumn{1}{c|}{65.0} & \multicolumn{1}{c|}{68.1}          & 69.5                                              &  & \multicolumn{1}{c|}{}                                                                       & \multicolumn{1}{c|}{38.4}          & \multicolumn{1}{c|}{\textbf{64.8}} & \multicolumn{1}{c|}{57.7}                                              & \multicolumn{1}{c|}{}                                                                       & \multicolumn{1}{c|}{57.0} & \multicolumn{1}{c|}{52.0} & 59.5                                              \\ \cline{1-11} \cline{13-20} \cline{22-29} 
\multirow{6}{*}{\begin{tabular}[c]{@{}c@{}}GPT-2\\ Large\end{tabular}} & \multicolumn{1}{c|}{B}                  &                                    & \multicolumn{1}{c|}{34.0}                                                                   & \multicolumn{1}{c|}{}              & \multicolumn{1}{c|}{}              & \multicolumn{1}{c|}{}                                                  & \multicolumn{1}{c|}{44.5}                                                                   & \multicolumn{1}{c|}{}              & \multicolumn{1}{c|}{}     &                                                   &  & \multicolumn{1}{c|}{27.1}                                                                   & \multicolumn{1}{c|}{}     & \multicolumn{1}{c|}{}              & \multicolumn{1}{c|}{}                                                  & \multicolumn{1}{c|}{42.3}                                                                   & \multicolumn{1}{c|}{}     & \multicolumn{1}{c|}{}              &                                                   &  & \multicolumn{1}{c|}{27.7}                                                                   & \multicolumn{1}{c|}{}              & \multicolumn{1}{c|}{}              & \multicolumn{1}{c|}{}                                                  & \multicolumn{1}{c|}{26.9}                                                                   & \multicolumn{1}{c|}{}     & \multicolumn{1}{c|}{}     &                                                   \\ \cline{2-11} \cline{13-20} \cline{22-29} 
                                                                       & \multicolumn{1}{c|}{AS}                  &                                    & \multicolumn{1}{c|}{}                                                                       & \multicolumn{1}{c|}{\textbf{63.2}} & \multicolumn{1}{c|}{54.8}          & \multicolumn{1}{c|}{-}                                                 & \multicolumn{1}{c|}{}                                                                       & \multicolumn{1}{c|}{\underline{54.7}}          & \multicolumn{1}{c|}{45.1} & -                                                 &  & \multicolumn{1}{c|}{}                                                                       & \multicolumn{1}{c|}{54.5} & \multicolumn{1}{c|}{45.2}          & \multicolumn{1}{c|}{-}                                                 & \multicolumn{1}{c|}{}                                                                       & \multicolumn{1}{c|}{46.0} & \multicolumn{1}{c|}{43.7}          & -                                                 &  & \multicolumn{1}{c|}{}                                                                       & \multicolumn{1}{c|}{\textbf{58.4}} & \multicolumn{1}{c|}{44.9}          & \multicolumn{1}{c|}{-}                                                 & \multicolumn{1}{c|}{}                                                                       & \multicolumn{1}{c|}{32.8} & \multicolumn{1}{c|}{\underline{33.6}} & -                                                 \\ \cline{2-11} \cline{13-20} \cline{22-29} 
                                                                       & \multicolumn{1}{c|}{\multirow{4}{*}{MP}} & $\star$                            & \multicolumn{1}{c|}{}                                                                       & \multicolumn{1}{c|}{22.1}          & \multicolumn{1}{c|}{44.9}          & \multicolumn{1}{c|}{30.5}                                              & \multicolumn{1}{c|}{}                                                                       & \multicolumn{1}{c|}{43.4}          & \multicolumn{1}{c|}{42.2} & 41.2                                              &  & \multicolumn{1}{c|}{}                                                                       & \multicolumn{1}{c|}{23.7} & \multicolumn{1}{c|}{30.1}          & \multicolumn{1}{c|}{39.2}                                              & \multicolumn{1}{c|}{}                                                                       & \multicolumn{1}{c|}{39.9} & \multicolumn{1}{c|}{43.8}          & 44.0                                              &  & \multicolumn{1}{c|}{}                                                                       & \multicolumn{1}{c|}{22.6}          & \multicolumn{1}{c|}{30.0}          & \multicolumn{1}{c|}{38.0}                                              & \multicolumn{1}{c|}{}                                                                       & \multicolumn{1}{c|}{25.2} & \multicolumn{1}{c|}{27.8} & 27.7                                              \\ \cline{3-11} \cline{13-20} \cline{22-29} 
                                                                       & \multicolumn{1}{c|}{}                    & "                                  & \multicolumn{1}{c|}{}                                                                       & \multicolumn{1}{c|}{29.7}          & \multicolumn{1}{c|}{41.2}          & \multicolumn{1}{c|}{41.8}                                              & \multicolumn{1}{c|}{}                                                                       & \multicolumn{1}{c|}{40.0}          & \multicolumn{1}{c|}{40.9} & 44.0                                              &  & \multicolumn{1}{c|}{}                                                                       & \multicolumn{1}{c|}{27.3} & \multicolumn{1}{c|}{31.3}          & \multicolumn{1}{c|}{36.8}                                              & \multicolumn{1}{c|}{}                                                                       & \multicolumn{1}{c|}{42.5} & \multicolumn{1}{c|}{47.3}          & \underline{49.1}                                              &  & \multicolumn{1}{c|}{}                                                                       & \multicolumn{1}{c|}{27.9}          & \multicolumn{1}{c|}{30.4}          & \multicolumn{1}{c|}{32.2}                                              & \multicolumn{1}{c|}{}                                                                       & \multicolumn{1}{c|}{27.7} & \multicolumn{1}{c|}{32.3} & 30.0                                              \\ \cline{3-11} \cline{13-20} \cline{22-29} 
                                                                       & \multicolumn{1}{c|}{}                    & <mark>      & \multicolumn{1}{c|}{}                                                                       & \multicolumn{1}{c|}{35.4}          & \multicolumn{1}{c|}{46.1}          & \multicolumn{1}{c|}{34.1}                                              & \multicolumn{1}{c|}{}                                                                       & \multicolumn{1}{c|}{35.6}          & \multicolumn{1}{c|}{45.8} & 25.1                                              &  & \multicolumn{1}{c|}{}                                                                       & \multicolumn{1}{c|}{25.6} & \multicolumn{1}{c|}{\textbf{56.5}} & \multicolumn{1}{c|}{51.1}                                              & \multicolumn{1}{c|}{}                                                                       & \multicolumn{1}{c|}{36.1} & \multicolumn{1}{c|}{48.4}          & 42.0                                              &  & \multicolumn{1}{c|}{}                                                                       & \multicolumn{1}{c|}{26.0}          & \multicolumn{1}{c|}{57.5}          & \multicolumn{1}{c|}{50.5}                                              & \multicolumn{1}{c|}{}                                                                       & \multicolumn{1}{c|}{22.5} & \multicolumn{1}{c|}{31.6} & 27.4                                              \\ \cline{3-11} \cline{13-20} \cline{22-29} 
                                                                       & \multicolumn{1}{c|}{}                    & <emphasize> & \multicolumn{1}{c|}{}                                                                       & \multicolumn{1}{c|}{34.8}          & \multicolumn{1}{c|}{46.7}          & \multicolumn{1}{c|}{45.4}                                              & \multicolumn{1}{c|}{}                                                                       & \multicolumn{1}{c|}{38.2}          & \multicolumn{1}{c|}{45.8} & 30.3                                              &  & \multicolumn{1}{c|}{}                                                                       & \multicolumn{1}{c|}{26.3} & \multicolumn{1}{c|}{52.2}          & \multicolumn{1}{c|}{55.1}                                              & \multicolumn{1}{c|}{}                                                                       & \multicolumn{1}{c|}{40.8} & \multicolumn{1}{c|}{47.7}          & 44.3                                              &  & \multicolumn{1}{c|}{}                                                                       & \multicolumn{1}{c|}{25.4}          & \multicolumn{1}{c|}{51.1}          & \multicolumn{1}{c|}{55.6}                                              & \multicolumn{1}{c|}{}                                                                       & \multicolumn{1}{c|}{25.4} & \multicolumn{1}{c|}{30.6} & 27.0                                              \\ \cline{1-11} \cline{13-20} \cline{22-29} 
\multirow{6}{*}{\begin{tabular}[c]{@{}c@{}}GPT-2\\ XL\end{tabular}}    & \multicolumn{1}{c|}{B}                  &                                    & \multicolumn{1}{c|}{28.0}                                                                   & \multicolumn{1}{c|}{}              & \multicolumn{1}{c|}{}              & \multicolumn{1}{c|}{}                                                  & \multicolumn{1}{c|}{51.2}                                                                   & \multicolumn{1}{c|}{}              & \multicolumn{1}{c|}{}     &                                                   &  & \multicolumn{1}{c|}{20.5}                                                                   & \multicolumn{1}{c|}{}     & \multicolumn{1}{c|}{}              & \multicolumn{1}{c|}{}                                                  & \multicolumn{1}{c|}{50.1}                                                                   & \multicolumn{1}{c|}{}     & \multicolumn{1}{c|}{}              &                                                   &  & \multicolumn{1}{c|}{24.8}                                                                   & \multicolumn{1}{c|}{}              & \multicolumn{1}{c|}{}              & \multicolumn{1}{c|}{}                                                  & \multicolumn{1}{c|}{31.8}                                                                   & \multicolumn{1}{c|}{}     & \multicolumn{1}{c|}{}     &                                                   \\ \cline{2-11} \cline{13-20} \cline{22-29} 
                                                                       & \multicolumn{1}{c|}{AS}                  &                                    & \multicolumn{1}{c|}{}                                                                       & \multicolumn{1}{c|}{34.0}          & \multicolumn{1}{c|}{39.9}          & \multicolumn{1}{c|}{-}                                                 & \multicolumn{1}{c|}{}                                                                       & \multicolumn{1}{c|}{\textbf{55.9}}          & \multicolumn{1}{c|}{45.7} & -                                                 &  & \multicolumn{1}{c|}{}                                                                       & \multicolumn{1}{c|}{35.5} & \multicolumn{1}{c|}{25.6}          & \multicolumn{1}{c|}{-}                                                 & \multicolumn{1}{c|}{}                                                                       & \multicolumn{1}{c|}{52.5} & \multicolumn{1}{c|}{52.4}          & -                                                 &  & \multicolumn{1}{c|}{}                                                                       & \multicolumn{1}{c|}{33.9}          & \multicolumn{1}{c|}{34.9}          & \multicolumn{1}{c|}{-}                                                 & \multicolumn{1}{c|}{}                                                                       & \multicolumn{1}{c|}{\underline{36.3}} & \multicolumn{1}{c|}{34.6} & -                                                 \\ \cline{2-11} \cline{13-20} \cline{22-29} 
                                                                       & \multicolumn{1}{c|}{\multirow{4}{*}{MP}} & $\star$                            & \multicolumn{1}{c|}{}                                                                       & \multicolumn{1}{c|}{28.9}          & \multicolumn{1}{c|}{31.0}          & \multicolumn{1}{c|}{41.7}                                              & \multicolumn{1}{c|}{}                                                                       & \multicolumn{1}{c|}{48.7}          & \multicolumn{1}{c|}{48.1} & 49.3                                              &  & \multicolumn{1}{c|}{}                                                                       & \multicolumn{1}{c|}{21.1} & \multicolumn{1}{c|}{25.7}          & \multicolumn{1}{c|}{32.1}                                              & \multicolumn{1}{c|}{}                                                                       & \multicolumn{1}{c|}{49.5} & \multicolumn{1}{c|}{51.2}          & 50.2                                              &  & \multicolumn{1}{c|}{}                                                                       & \multicolumn{1}{c|}{23.2}          & \multicolumn{1}{c|}{27.2}          & \multicolumn{1}{c|}{28.1}                                              & \multicolumn{1}{c|}{}                                                                       & \multicolumn{1}{c|}{31.8} & \multicolumn{1}{c|}{34.0} & 33.8                                              \\ \cline{3-11} \cline{13-20} \cline{22-29} 
                                                                       & \multicolumn{1}{c|}{}                    & "                                  & \multicolumn{1}{c|}{}                                                                       & \multicolumn{1}{c|}{30.2}          & \multicolumn{1}{c|}{35.8}          & \multicolumn{1}{c|}{43.7}                                              & \multicolumn{1}{c|}{}                                                                       & \multicolumn{1}{c|}{50.0}          & \multicolumn{1}{c|}{46.0} & 46.1                                              &  & \multicolumn{1}{c|}{}                                                                       & \multicolumn{1}{c|}{23.5} & \multicolumn{1}{c|}{29.9}          & \multicolumn{1}{c|}{37.5}                                              & \multicolumn{1}{c|}{}                                                                       & \multicolumn{1}{c|}{49.8} & \multicolumn{1}{c|}{51.8}          & 51.9                                              &  & \multicolumn{1}{c|}{}                                                                       & \multicolumn{1}{c|}{25.6}          & \multicolumn{1}{c|}{28.2}          & \multicolumn{1}{c|}{30.7}                                              & \multicolumn{1}{c|}{}                                                                       & \multicolumn{1}{c|}{32.2} & \multicolumn{1}{c|}{33.6} & 33.6                                              \\ \cline{3-11} \cline{13-20} \cline{22-29} 
                                                                       & \multicolumn{1}{c|}{}                    & <mark>      & \multicolumn{1}{c|}{}                                                                       & \multicolumn{1}{c|}{30.1}          & \multicolumn{1}{c|}{43.3}          & \multicolumn{1}{c|}{\underline{51.0}}                                     & \multicolumn{1}{c|}{\textbf{}}                                                              & \multicolumn{1}{c|}{49.8}          & \multicolumn{1}{c|}{49.5} & 47.0                                              &  & \multicolumn{1}{c|}{}                                                                       & \multicolumn{1}{c|}{17.4} & \multicolumn{1}{c|}{\underline{38.3}}          & \multicolumn{1}{c|}{36.2}                                              & \multicolumn{1}{c|}{}                                                                       & \multicolumn{1}{c|}{47.3} & \multicolumn{1}{c|}{\textbf{53.4}} & 49.9                                              &  & \multicolumn{1}{c|}{}                                                                       & \multicolumn{1}{c|}{19.0}          & \multicolumn{1}{c|}{\textbf{38.1}} & \multicolumn{1}{c|}{34.8}                                              & \multicolumn{1}{c|}{}                                                                       & \multicolumn{1}{c|}{29.8} & \multicolumn{1}{c|}{34.8} & 31.6                                              \\ \cline{3-11} \cline{13-20} \cline{22-29} 
                                                                       & \multicolumn{1}{c|}{}                    & <emphasize> & \multicolumn{1}{c|}{}                                                                       & \multicolumn{1}{c|}{28.4}          & \multicolumn{1}{c|}{42.3}          & \multicolumn{1}{c|}{42.9}                                              & \multicolumn{1}{c|}{}                                                                       & \multicolumn{1}{c|}{48.2}          & \multicolumn{1}{c|}{50.4} & 46.1                                              &  & \multicolumn{1}{c|}{}                                                                       & \multicolumn{1}{c|}{18.2} & \multicolumn{1}{c|}{32.3}          & \multicolumn{1}{c|}{37.9}                                              & \multicolumn{1}{c|}{}                                                                       & \multicolumn{1}{c|}{48.7} & \multicolumn{1}{c|}{\textbf{53.4}} & 50.4                                              &  & \multicolumn{1}{c|}{}                                                                       & \multicolumn{1}{c|}{20.8}          & \multicolumn{1}{c|}{32.1}          & \multicolumn{1}{c|}{34.2}                                              & \multicolumn{1}{c|}{}                                                                       & \multicolumn{1}{c|}{30.0} & \multicolumn{1}{c|}{35.2} & 32.4                                              \\ \cline{1-11} \cline{13-20} \cline{22-29} 
\end{tabular}
\caption{Question vs. Context Table: B=Baseline (no emphasis); AS=Attention steering; MP=Marked prompting; C=Context; Q=Question; <q>=question string; <c>=context string; The highest score for each model is in bold, the second highest on the other prompt structure is underlined. The AS method requires a substring within the input string to be emphasized, and hence, it is undefined for the Q+C setting, as in that setting the substring will be the entire input string.}
\label{question_vs_context}
\end{table*}
%%%%%

We first analyze whether models' performance differs when given the same information, but in different order: question-first and context-first.
% once when the question is first in the prompt, and once when the context is first. 
Our results can be seen in Table \ref{question_vs_context}. 

\paragraph{No Emphasis Accuracy}
As we aim to understand the effect that prompt structure alone has on models' performance, for this analysis we focus on the no emphasis (NE) baseline.
% as the different emphasis methods might skew the result. 
% However, in Section \ref{PS+Emph}, we analyze the effect that prompt structure and prompt emphasis combined have on models' performance. 

Looking at the NE setting, there is a clear difference across almost all models and datasets. More specifically, prompting models with the context first strongly improves performance, with an average increase of $13.46\%$ ($49.90\%$ in comparison to $36.44\%$). On the Natural Questions dataset, the highest accuracy change occurs for GPT-J: from $33.3\%$ to $64.5\%$ ($31.2\%$ difference). The second highest change is seen for GPT-2-XL: from $28.0\%$ to $51.2\%$ ($23.2\%$ difference). The third highest change occurs for Llama-2, which scores $46.3\%$ when the question is given first but $58.1\%$ when the context is given first ($11.8\%$ difference). Similar behavior can be seen for the SQuAD and AdversarialQA datasets as well. For example, Llama-2 changes from $60.4\%$ to $72.9\%$ on SQuAD, and GPT-2-XL changes from $24.8\%$ to $31.8\%$ on AdversarialQA. However, we do find two cases where placing the context first does not improve the results, and actually slightly reduces them: on the AdversarialQA dataset, GPT-2 large and GPT-J change from $27.7\%$ to $26.9\%$ and $47.2\%$ to $46.2\%$, respectively.

% \subsection{RQ 2: Comparing Emphasis Strategies}
\subsection{RQ 2: Emphasis and Performance}
We next analyze whether emphasizing parts of the input -- the question, the context, or both -- enhances models' performance. Our results can be seen again in Table \ref{question_vs_context}.

\paragraph{Performance Improvement Across Almost All Settings}
We find that across all datasets, models, and prompt structures, there is a performance difference between emphasizing either the context, the question, or both, which will further be discussed in Section \ref{c_q_cq}. However, emphasizing parts of the input is overall beneficial and can strongly improve models' NE performance. For example, on the Natural Questions dataset, every emphasis method improves Llama-2 NE performance for the question-first setting (except for emphasizing the context using MP-*). To more concretely assess the overall performance improvement emphasizing the input entails, we compare the averaged NE performance across all models, dataset, and settings, to the averaged performance over all emphasis methods, models, datasets, and settings. We find that, while the average NE performance is $43.17\%$, the average model performance when emphasizing the input is $47.31\%$. 

\section{Analysis and Discussion}

\subsection{Sequence Order Analysis}
\paragraph{No-emphasis Perplexity}
%%%%%
\begin{table}[]
\centering
\small
\setlength{\tabcolsep}{0.7pt}
\begin{tabular}{|c|cc|c|cc|c|cc|}
\cline{1-3} \cline{5-6} \cline{8-9}
\textbf{Model}                                           & \multicolumn{2}{c|}{\textbf{NQ}}                                                                                                                      &  & \multicolumn{2}{c|}{\textbf{SQuAD}}                                                                                                                                  &  & \multicolumn{2}{c|}{\textbf{AdversarialQA}}                                                                                                                          \\ \cline{1-3} \cline{5-6} \cline{8-9} 
                                                         & \multicolumn{1}{c|}{\begin{tabular}[c]{@{}c@{}}Question\\ First\end{tabular}} & {\begin{tabular}[c]{@{}c@{}}Context\\ First\end{tabular}}{} &  & \multicolumn{1}{c|}{\begin{tabular}[c]{@{}c@{}}Question\\ First\end{tabular}} & {\begin{tabular}[c]{@{}c@{}}Context\\ First\end{tabular}}{} &  & \multicolumn{1}{c|}{\begin{tabular}[c]{@{}c@{}}Question\\ First\end{tabular}} & {\begin{tabular}[c]{@{}c@{}}Context\\ First\end{tabular}}{} \\ \cline{1-3} \cline{5-6} \cline{8-9} 
Llama                                                 & \multicolumn{1}{c|}{\textbf{15.08}}                                                         & 15.53                                                                  &  & \multicolumn{1}{c|}{11.49}                                                                  & \textbf{10.58}                                                         &  & \multicolumn{1}{c|}{12.89}                                                                  & \textbf{11.96}                                                         \\ \cline{1-3} \cline{5-6} \cline{8-9} 
GPT-J                                                    & \multicolumn{1}{c|}{20.16}                                                         & \textbf{18.61}                                                                  &  & \multicolumn{1}{c|}{13.13}                                                                  & \textbf{13.07}                                                         &  & \multicolumn{1}{c|}{14.52}                                                                  & \textbf{14.36}                                                         \\ \cline{1-3} \cline{5-6} \cline{8-9} 
\begin{tabular}[c]{@{}c@{}}GPT-2 \\ Large\end{tabular} & \multicolumn{1}{c|}{36.26}                                                                  & \textbf{32.22}                                                         &  & \multicolumn{1}{c|}{\textbf{20.86}}                                                         & 21.24                                                                  &  & \multicolumn{1}{c|}{\textbf{22.88}}                                                         & 23.29                                                                  \\ \cline{1-3} \cline{5-6} \cline{8-9} 
\begin{tabular}[c]{@{}c@{}}GPT-2\\ XL\end{tabular}    & \multicolumn{1}{c|}{30.44}                                                                  & \textbf{28.47}                                                         &  & \multicolumn{1}{c|}{19.02}                                                                  & \textbf{18.89}                                                         &  & \multicolumn{1}{c|}{20.99}                                                                  & \textbf{20.70}                                                         \\ \cline{1-3} \cline{5-6} \cline{8-9} 
\end{tabular}
\caption{Model’s average perplexity on each dataset, for each prompt structure, in the zero shot (no emphasis) setting. Lower is better. NQ=Natural Questions.}
\label{ppl_table}
\end{table}
%%%%%
To further understand the behavior we find from our analysis of RQ1 in Section \ref{rq1_results}, we evaluate the average perplexity of the prompts under each model for each of the two prompt structures -- \textit{Question: <q> Context: <c>} and \textit{Context: <c> Question: <q>} --, each dataset and the NE setting. Our results can be seen in Table \ref{ppl_table}. 

Across almost all dataset, models' perplexity is lower (i.e., “better”) for the context-first setting, with an average reduction of $1.77$ on the Natural Questions ($25.48$ vs. $23.70$), $1.77$ on SQuAD ($16.12$ vs. $15.94$), $0.24$ on AdversarialQA ($17.82$ vs. $17.57$), and over all datasets of $0.73$ ($19.81$ vs. $19.07$). For example, the highest perplexity reduction occurs for GPT-2 large, which scores $32.22$ on the Natural Questions dataset when the context is provided first, in comparison to $36.26$ for the question-first setting ($4.04$ difference). 

\paragraph{Perplexity vs. Accuracy}
Surprisingly, looking at Table \ref{ppl_table} for the two cases above in which placing the context first does not improve accuracy (GPT-2 large and GPT-J on AdversarialQA), we find that only GPT-2 large scores higher on perplexity for the context-first setting, which could potentially explain the accuracy difference as the model finds this prompt structure more confusing on this particular dataset. However, we do not find that the perplexity was higher for the questions-first structure for GPT-J. Moreover, we find two more cases where models' perplexity was higher for one of the structures, but accuracy was higher on the same structure: Llama-2 on Natural Questions and GPT-2 large on SQuAD. This suggests that while the models do not find the context-first structure more confusing (as measured by their perplexity), they score lower on accuracy for another reason.

\subsection{Emphasis Analysis}
\paragraph{Different Emphasis Methods Affect Similar Models Differently}
We find that different emphasis methods affect similar models differently. On the Natural Questions dataset, while emphasizing the context using the MP-<emphasize> method on GPT-J on the question-first structure increases its NE accuracy from $33.3\%$ to $69.0\%$, outperforming all other models, using the MP-* method reduces its score to $26.9\%$. 
% Similar behavior can be seen on other datasets and models. For example, on the SQuAD dataset, GPT-2 large increase from $42.3\%$ to $49.1\%$ when emphasizing the question and context using the MP-" method on the context-first structure, but decreases to $42.0\%$ when using the MP-<mark> method.

% We can also see that the second highest increase occur for GPT-2 large, where on the same dataset and setting its score change from $34.0\%$ to $63.2\%$ when applying AS to the question string, resulting in it outperforming Llama-2, which has 10 times more parameters.

\paragraph{Similar Emphasis Methods Affect Different Models Differently}
We also find that similar emphasis methods affect different models differently. For example, on the AdversarialQA dataset and the context-first, context-emphasis setting, AS improves Llama-2 NE performance from $49.4\%$ to $53.3\%$, and GPT-2-XL's NE performance from $26.9\%$ to $33.6\%$. However, AS reduces GPT-J's performance from $46.2\%$ to $41.7\%$. 

\paragraph{Best Emphasis Methods}
To assess which emphasis methods are best for each model, we average the scores across all datasets and settings for each model. We find that the top 3 best emphasis methods for each model are (in decreasing order): Llama-2: (MP-", MP-<mark>, MP-<emphasize>), GPT-J: (MP-<emphasize>, MP-<mark>, MP-"), GPT-2 large: (AS, MP-<emphasize>, MP-<mark>), and GPT-2-XL: (AS, MP-<mark>, MP-<emphasize>). 

Overall, across all models, datasets and settings, the best emphasis method may seem to be AS, with an average accuracy of $49.39\%$. This is aligned with \citet{zhang2023tell}'s result, which finds that AS outperforms two MP methods on the task of instruction following.

However, looking at the top accuracies for each model on each dataset, we actually find that AS only outperforms other emphasis methods 6 out of the 24 times (4 models, 2 prompt structures for each, on 3 datasets). And from that regard, MP outperforms it (MP also scores fairly close to it overall, with the highest average accuracy of $48.68\%$ for MP-<emphasize>). 
% This is surprising, as it only requires a concatenation of a few tokens to the input.
% , in comparison to the AS method, which is far more involved. 

\paragraph{Emphasis on C vs. Q vs. CQ}
\label{c_q_cq}
To analyze which substring is better to emphasize -- the context, the question, or both --, we average the performance of all models across all datasets, emphasis methods, and prompt structures. We find that the highest performance is achieved by emphasizing both context and question, with an average accuracy score of $49.49\%$. However, we also find that emphasizing the context is roughly just as good, with an average accuracy score of $49.21\%$, and that emphasizing the question falls much below both, with an average accuracy score of $43.68\%$.

\paragraph{Does Size Matter?}
Here, we analyze whether models' size affects their ability to be emphasized by looking at the best method for each on each setting. And while we do not find a clear pattern, we find some cases that suggest that emphasis methods are more beneficial for smaller models. For example, on the SQuAD dataset and the question-first setting, GPT-2 large improves from $27.1\%$ to $56.5\%$ using the MP-<mark> method ($29.4\%$ improvement), 
% where GPT-2-XL improve from $20.5\%$ to $38.3\%$ using the MP-<mark> method ($17.8\%$ improvement), 
where GPT-J improves from $45.5\%$ to $64.7\%$ using the MP-<emphasis> method ($19.2\%$ improvement), and Llama-2 from $60.4\%$ to $72.3\%$ using the MP-" method ($11.9\%$ improvement).

\paragraph{Does Training Data Matter?}
To evaluate the effect training data has on the susceptibility of models for being emphasized, we compare GPT-2 large and GPT-2-XL as they are trained on the same corpus. From Table \ref{question_vs_context} we can see that, while these two models are trained on similar data, on many occasions, similar emphasis methods result in different behavior. For example, on the question-first setting and the Natural Questions dataset, while AS result in the highest performance when applied to the question on both models, for context emphasis, the best method for GPT-2 large is AS, where for GPT-2-XL the best method is MP-<mark> or MP-<emphasize>. We also do not find the same absolute improvements across the two models when looking at similar emphasis methods and similar settings. This suggests that, while the training data has some effect on which emphasis method is beneficial for each model, it is not the whole story.

\paragraph{Attention Heads Analysis}

\begin{table}[]
\centering
\tiny
\setlength{\tabcolsep}{1.0pt}
\begin{tabular}{|c|c|cc|cc|}
\hline
\textbf{Model}                                                         & \textbf{Emphasis Method} & \multicolumn{2}{c|}{\textbf{\begin{tabular}[c]{@{}c@{}}Question\\ Emphasis\end{tabular}}}                        & \multicolumn{2}{c|}{\textbf{\begin{tabular}[c]{@{}c@{}}Context\\ Emphasis\end{tabular}}}                        \\ \hline
                                                                       &                          & \multicolumn{1}{c|}{Accuracy} & \begin{tabular}[c]{@{}c@{}}Question String\\ Avg. Attention\\ Score\end{tabular} & \multicolumn{1}{c|}{Accuracy} & \begin{tabular}[c]{@{}c@{}}Context String\\ Avg. Attention\\ Score\end{tabular} \\ \hline
\multirow{4}{*}{\begin{tabular}[c]{@{}c@{}}GPT 2\\ Large\end{tabular}} & *                        & \multicolumn{1}{c|}{22.1}     & 0.0078                                                                           & \multicolumn{1}{c|}{44.9}     & 0.0041                                                                          \\ \cline{2-6} 
                                                                       & "                        & \multicolumn{1}{c|}{29.7}     & 0.0078                                                                           & \multicolumn{1}{c|}{41.2}     & 0.0094                                                                          \\ \cline{2-6} 
                                                                       & mark                     & \multicolumn{1}{c|}{35.4}     & 0.0074                                                                           & \multicolumn{1}{c|}{46.1}     & 0.0088                                                                          \\ \cline{2-6} 
                                                                       & emphasis                 & \multicolumn{1}{c|}{34.8}     & 0.0070                                                                           & \multicolumn{1}{c|}{46.7}     & 0.0084                                                                          \\ \hline
\multirow{4}{*}{\begin{tabular}[c]{@{}c@{}}GPT 2\\ XL\end{tabular}}    & *                        & \multicolumn{1}{c|}{28.9}     & 0.0076                                                                           & \multicolumn{1}{c|}{31.0}     & 0.0039                                                                          \\ \cline{2-6} 
                                                                       & "                        & \multicolumn{1}{c|}{30.2}     & 0.0075                                                                           & \multicolumn{1}{c|}{35.8}     & 0.0095                                                                          \\ \cline{2-6} 
                                                                       & mark                     & \multicolumn{1}{c|}{30.1}     & 0.0071                                                                           & \multicolumn{1}{c|}{43.3}     & 0.0089                                                                          \\ \cline{2-6} 
                                                                       & emphasis                 & \multicolumn{1}{c|}{28.4}     & 0.0067                                                                           & \multicolumn{1}{c|}{42.3}     & 0.0085                                                                          \\ \hline
\end{tabular}
\caption{Attention scores analysis across different models' layers and heads for different emphasis methods.}
\label{attention_heads_table}
\end{table}

To further understand why different emphasis methods result in different models' scores we evaluate the attention scores for the strings that are being emphasized by the different methods on the question-first setting. More concretely, for each MP method, we send each sentence from the Natural Questions dataset to the model. We then average the attention scores across all model's heads and layers for the tokens corresponding to the string to be emphasized -- either the context or the question. Our results can be seen in Table \ref{attention_heads_table}. 

We do not find a clear pattern that highlights whether emphasis methods result in a higher or lower attention scores for emphasis strings. For example, while GPT 2 large has an increase of accuracy from $22.1\%$ to $29.7\%$ when changing from the MP-* method to the MP-" method on the question-emphasis setting, the attention scores stay the same. We also see that sometimes the attention scores go up when accuracy go down, such as in GPT 2 XL, MP-mark to MP-* on question emphasis, and sometimes the attention scores go down when accuracy go up, such as in GPT 2 large, MP-" to MP-emphasis, on the context emphasis setting.

\subsection{Known Vs. Unknown Knowledge}

\paragraph{Marked Prompting}
\begin{table*}[]
\centering
\tiny
\setlength{\tabcolsep}{0.6pt}\begin{tabular}{|c|ccccc|c|ccccc|c|ccccc|}
\cline{1-6} \cline{8-12} \cline{14-18}
\textbf{Model}                                        & \multicolumn{5}{c|}{\textbf{Natural Questions}}                                                                                                                                                                                                                                                                                                                                                      &  & \multicolumn{5}{c|}{\textbf{SQuAD}}                                                                                                                                                                                                                                                                                                                                                                  &  & \multicolumn{5}{c|}{\textbf{AdversarialQA}}                                                                                                                                                                                                                                                                                                                                                          \\ \cline{1-6} \cline{8-12} \cline{14-18} 
                                                      & \multicolumn{1}{c|}{\begin{tabular}[c]{@{}c@{}}Knowledge\\ Amount\end{tabular}} & \multicolumn{1}{c|}{\begin{tabular}[c]{@{}c@{}}Known\\ No Emphasis\end{tabular}} & \multicolumn{1}{c|}{\begin{tabular}[c]{@{}c@{}}Known\\ Emphasis\end{tabular}} & \multicolumn{1}{c|}{\begin{tabular}[c]{@{}c@{}}Unknown\\ No Emphasis\end{tabular}} & \begin{tabular}[c]{@{}c@{}}Unknown\\ Emphasis\end{tabular} &  & \multicolumn{1}{c|}{\begin{tabular}[c]{@{}c@{}}Knowledge\\ Amount\end{tabular}} & \multicolumn{1}{c|}{\begin{tabular}[c]{@{}c@{}}Known\\ No Emphasis\end{tabular}} & \multicolumn{1}{c|}{\begin{tabular}[c]{@{}c@{}}Known\\ Emphasis\end{tabular}} & \multicolumn{1}{c|}{\begin{tabular}[c]{@{}c@{}}Unknown\\ No Emphasis\end{tabular}} & \begin{tabular}[c]{@{}c@{}}Unknown\\ Emphasis\end{tabular} &  & \multicolumn{1}{c|}{\begin{tabular}[c]{@{}c@{}}Knowledge\\ Amount\end{tabular}} & \multicolumn{1}{c|}{\begin{tabular}[c]{@{}c@{}}Known\\ No Emphasis\end{tabular}} & \multicolumn{1}{c|}{\begin{tabular}[c]{@{}c@{}}Known\\ Emphasis\end{tabular}} & \multicolumn{1}{c|}{\begin{tabular}[c]{@{}c@{}}Unknown\\ No Emphasis\end{tabular}} & \begin{tabular}[c]{@{}c@{}}Unknown\\ Emphasis\end{tabular} \\ \cline{1-6} \cline{8-12} \cline{14-18} 
Llama 2                                               & \multicolumn{1}{c|}{20.0}                                                       & \multicolumn{1}{c|}{93.4}                                                        & \multicolumn{1}{c|}{93.6}                                                     & \multicolumn{1}{c|}{\textbf{46.4}}                                                 & \textbf{49.9}                                              &  & \multicolumn{1}{c|}{18.1}                                                       & \multicolumn{1}{c|}{88.6}                                                        & \multicolumn{1}{c|}{91.9}                                                     & \multicolumn{1}{c|}{\textbf{70.0}}                                                 & \textbf{79.7}                                              &  & \multicolumn{1}{c|}{20.5}                                                       & \multicolumn{1}{c|}{77.9}                                                        & \multicolumn{1}{c|}{71.7}                                                     & \multicolumn{1}{c|}{\textbf{42.7}}                                                 & \textbf{51.9}                                              \\ \cline{1-6} \cline{8-12} \cline{14-18} 
GPT J                                                 & \multicolumn{1}{c|}{4.3}                                                        & \multicolumn{1}{c|}{90.2}                                                        & \multicolumn{1}{c|}{89.5}                                                     & \multicolumn{1}{c|}{\textbf{63.2}}                                                 & \textbf{65.5}                                              &  & \multicolumn{1}{c|}{9.2}                                                        & \multicolumn{1}{c|}{83.5}                                                        & \multicolumn{1}{c|}{86.5}                                                     & \multicolumn{1}{c|}{\textbf{58.7}}                                                 & \textbf{71.2}                                              &  & \multicolumn{1}{c|}{14.2}                                                       & \multicolumn{1}{c|}{71.3}                                                        & \multicolumn{1}{c|}{73.7}                                                     & \multicolumn{1}{c|}{\textbf{42.0}}                                                 & \textbf{58.0}                                              \\ \cline{1-6} \cline{8-12} \cline{14-18} 
\begin{tabular}[c]{@{}c@{}}GPT 2\\ Large\end{tabular} & \multicolumn{1}{c|}{1.7}                                                        & \multicolumn{1}{c|}{\textbf{78.8}}                                               & \multicolumn{1}{c|}{\textbf{86.4}}                                            & \multicolumn{1}{c|}{43.7}                                                          & 43.1                                                       &  & \multicolumn{1}{c|}{4.6}                                                        & \multicolumn{1}{c|}{79.6}                                                        & \multicolumn{1}{c|}{84.1}                                                     & \multicolumn{1}{c|}{\textbf{40.5}}                                                 & \textbf{47.6}                                              &  & \multicolumn{1}{c|}{11.4}                                                       & \multicolumn{1}{c|}{64.9}                                                        & \multicolumn{1}{c|}{61.9}                                                     & \multicolumn{1}{c|}{\textbf{22.0}}                                                 & \textbf{25.9}                                              \\ \cline{1-6} \cline{8-12} \cline{14-18} 
\begin{tabular}[c]{@{}c@{}}GPT 2\\ XL\end{tabular}    & \multicolumn{1}{c|}{2.2}                                                        & \multicolumn{1}{c|}{88.5}                                                        & \multicolumn{1}{c|}{85.2}                                                     & \multicolumn{1}{c|}{\textbf{50.2}}                                                 & \textbf{48.4}                                              &  & \multicolumn{1}{c|}{6.0}                                                        & \multicolumn{1}{c|}{78.5}                                                        & \multicolumn{1}{c|}{79.1}                                                     & \multicolumn{1}{c|}{\textbf{48.2}}                                                 & \textbf{50.3}                                              &  & \multicolumn{1}{c|}{11.9}                                                       & \multicolumn{1}{c|}{\textbf{67.5}}                                               & \multicolumn{1}{c|}{\textbf{69.8}}                                            & \multicolumn{1}{c|}{26.9}                                                          & 28.9                                                       \\ \cline{1-6} \cline{8-12} \cline{14-18} 
\end{tabular}
\caption{Known vs. Unknown Table: \textbf{Marked Prompting}. We find that the best emphasizing method is marked prompting, and in particular, concatenating the string “<emphasize>” before and after the context and question strings. We use the closed-book setting to evaluate models' parametric knowledge, and compare the ZS baseline (no emphasis) to the best marked prompting approach. In bold, the largest improvement for each model on each dataset. Knowledge Amount is measured using accuracy, as the average number of questions models can successfully answer correctly without context (cf. Section \ref{knowledge_amount}).}
\label{known_vs_unknown_mp}
\end{table*}
\begin{table*}[]
\centering
\tiny
\setlength{\tabcolsep}{1.0pt}
\begin{tabular}{|c|cccc|c|cccc|c|cccc|}
\cline{1-5} \cline{7-10} \cline{12-15}
\textbf{Model / Dataset} & \multicolumn{4}{c|}{\textbf{Natural Questions}}                                                                                                                                                                                                                                                               & \multicolumn{1}{l|}{} & \multicolumn{4}{c|}{\textbf{SQuAD}}                                                                                                                                                                                                                                                                           & \multicolumn{1}{l|}{} & \multicolumn{4}{c|}{\textbf{AdversarialQA}}                                                                                                                                                                                                                                                                   \\ \cline{1-5} \cline{7-10} \cline{12-15} 
                         & \multicolumn{1}{c|}{\begin{tabular}[c]{@{}c@{}}Known \\ No Emphasis\end{tabular}} & \multicolumn{1}{c|}{\begin{tabular}[c]{@{}c@{}}Known\\ Steering\end{tabular}} & \multicolumn{1}{c|}{\begin{tabular}[c]{@{}c@{}}Unknown\\ No Emphasis\end{tabular}} & \begin{tabular}[c]{@{}c@{}}Unknown\\ Steering\end{tabular} &                       & \multicolumn{1}{c|}{\begin{tabular}[c]{@{}c@{}}Known \\ No Emphasis\end{tabular}} & \multicolumn{1}{c|}{\begin{tabular}[c]{@{}c@{}}Known\\ Steering\end{tabular}} & \multicolumn{1}{c|}{\begin{tabular}[c]{@{}c@{}}Unknown\\ No Emphasis\end{tabular}} & \begin{tabular}[c]{@{}c@{}}Unknown\\ Steering\end{tabular} &                       & \multicolumn{1}{c|}{\begin{tabular}[c]{@{}c@{}}Known \\ No Emphasis\end{tabular}} & \multicolumn{1}{c|}{\begin{tabular}[c]{@{}c@{}}Known\\ Steering\end{tabular}} & \multicolumn{1}{c|}{\begin{tabular}[c]{@{}c@{}}Unknown\\ No Emphasis\end{tabular}} & \begin{tabular}[c]{@{}c@{}}Unknown\\ Steering\end{tabular} \\ \cline{1-5} \cline{7-10} \cline{12-15} 
Llama-2                  & \multicolumn{1}{c|}{\textbf{69.1}}                                             & \multicolumn{1}{c|}{\textbf{81.0}}                                            & \multicolumn{1}{c|}{30.8}                                                       & 37.0                                                       &                       & \multicolumn{1}{c|}{80.6}                                                      & \multicolumn{1}{c|}{85.8}                                                     & \multicolumn{1}{c|}{\textbf{56.7}}                                              & \textbf{62.0}                                              &                       & \multicolumn{1}{c|}{69.8}                                                      & \multicolumn{1}{c|}{67.8}                                                     & \multicolumn{1}{c|}{\textbf{38.0}}                                              & \textbf{38.5}                                              \\ \cline{1-5} \cline{7-10} \cline{12-15} 
GPT-J                    & \multicolumn{1}{c|}{56.4}                                                      & \multicolumn{1}{c|}{76.8}                                                     & \multicolumn{1}{c|}{\textbf{27.9}}                                              & \textbf{59.9}                                              &                       & \multicolumn{1}{c|}{53.4}                                                      & \multicolumn{1}{c|}{66.5}                                                     & \multicolumn{1}{c|}{\textbf{44.7}}                                              & \textbf{62.6}                                              &                       & \multicolumn{1}{c|}{\textbf{63.8}}                                             & \multicolumn{1}{c|}{\textbf{63.8}}                                            & \multicolumn{1}{c|}{44.4}                                                       & 41.9                                                       \\ \cline{1-5} \cline{7-10} \cline{12-15} 
GPT-2 Large              & \multicolumn{1}{c|}{52.1}                                                      & \multicolumn{1}{c|}{71.4}                                                     & \multicolumn{1}{c|}{\textbf{29.4}}                                              & \textbf{54.6}                                              &                       & \multicolumn{1}{c|}{47.1}                                                      & \multicolumn{1}{c|}{68.9}                                                     & \multicolumn{1}{c|}{\textbf{26.1}}                                              & \textbf{53.7}                                              &                       & \multicolumn{1}{c|}{42.1}                                                      & \multicolumn{1}{c|}{59.6}                                                     & \multicolumn{1}{c|}{\textbf{25.8}}                                              & \textbf{58.3}                                              \\ \cline{1-5} \cline{7-10} \cline{12-15} 
GPT-2 XL                 & \multicolumn{1}{c|}{51.4}                                                      & \multicolumn{1}{c|}{49.1}                                                     & \multicolumn{1}{c|}{\textbf{24.2}}                                              & \textbf{33.5}                                              &                       & \multicolumn{1}{c|}{39.1}                                                      & \multicolumn{1}{c|}{52.4}                                                     & \multicolumn{1}{c|}{\textbf{19.3}}                                              & \textbf{34.4}                                              &                       & \multicolumn{1}{c|}{\textbf{40.7}}                                             & \multicolumn{1}{c|}{\textbf{50.8}}                                            & \multicolumn{1}{c|}{20.9}                                                       & 29.6                                                       \\ \cline{1-5} \cline{7-10} \cline{12-15} 
\end{tabular}
\caption{Known vs. Unknown Table: \textbf{Attention Steering}. While attention steering does not overall perform as well as marked prompting, we also evaluate models' parametric knowledge (known vs. unknown) using the closed-book setting, and compare the ZS No Emphasis (no emphasis) to the attention steering approach where the question is presented first in the prompt and is being emphasized -- as that is the best setting we find for attention steering. In bold, the largest improvement for each model on each dataset.}
\label{known_vs_unknown_as}
\end{table*}
We next evaluate whether MP, and specifically the best performing setting overall -- context-first, question + context emphasis --, works better for addressing knowledge that models have or do not have. Our results can be seen in Table \ref{known_vs_unknown_mp}.

We can see that, across almost all three datasets and all models, emphasizing the input string on the unknown knowledge split results in more improvement than emphasizing the input string on the known knowledge split. For example, on Natural Questions, for unknown knowledge, Llama-2 and GPT-J improve from $46.4\%$ and $63.2\%$ to $49.9\%$ and $65.5\%$, respectively. Where on the known knowledge split, they respectively change from $93.4\%$ to $93.6\%$ and from $88.5\%$ to $85.2\%$. 

One potential explanation for that is that models tend to already perform reasonably well on known knowledge, since they have most likely acquired that knowledge during training. However, emphasizing input strings on unknown knowledge forces the model to adapt its learned representations to handle unseen or less familiar data.

% However, we find that on two occasions emphasizing the input string on the known knowledge split outperforms emphasizing the input on the unknown split: 1) On the Natural Questions dataset, GPT-2 large improves from $78.8\%$ to $86.4\%$ in the known knowledge split, in comparison to a decrease in performance from $43.7\%$ to $43.1\%$ in the unknown knowledge split. And 2) on AdversarialQA GPT-2-XL improves from $67.5\%$ to $69.8\%$ in the known knowledge split, in comparison to an improvement from $26.9\%$ to $28.9\%$ on the unknown knowledge split.

\paragraph{Attention Steering}
Next, we evaluate whether AS, and specifically the best performing setting of AS -- question-first, question steering --, works better for addressing knowledge that models have or do not have. Our results can be seen in Table \ref{known_vs_unknown_as}.

Across almost all three datasets and all models, steering the input string in the unknown knowledge split results in more improvement than steering it in the known knowledge split. For example, on Natural Questions, for unknown knowledge, GPT-J and GPT-2 Large improve from $27.9\%$ and $29.4\%$ to $59.9\%$ and $54.6\%$, respectively. In contrast, on the known knowledge split, they improve from $56.4\%$ to $76.8\%$ and from $52.1\%$ to $71.4\%$, respectively.

% However, on three occasions steering the input string in the known knowledge split outperform steering it in the unknown split. In particular, on Natural Questions Llama-2 improves from $69.1\%$ to $81.0\%$ (in comparison to an improve of $6.2\%$ for unknown knowledge), and on AdversarialQA GPT-J improves from $63.8\%$ to $63.8\%$ (in comparison to a decrease of $2.5\%$ for unknown knowledge), where GPT-2-XL improves from $40.7\%$ to $50.8\%$ (in comparison to an improve of $8.7\%$ for unknown knowledge).

% \subsection{Does the emphasis methods have anything to do with which sequence in first/last in the sequence? e.g., they work better on stuff that are in the beginning/end, regardless if its the question or context}

\subsection{Can Emphasis Be Bad?}
While we find that emphasizing parts of the input using various emphasis methods can be beneficial, it does require experimentation, as choosing the wrong emphasis method can actually be disadvantageous. Averaging over all datasets, models, and settings in Table \ref{question_vs_context}, we find that the worse emphasis method is MP-*, only increasing the average accuracy from $43.17\%$ to $43.66\%$, and at its worst setting it reduces Llama-2's baseline performance from $46.3\%$ to $31.6\%$ on the Natural Questions dataset in the question-first setting.

\subsection{Newer Models, Instruction Tuning, and Max Context Length}
\label{additional_models_sec}
\begin{table}[]
\centering
\tiny
\setlength{\tabcolsep}{0.6pt}
\begin{tabular}{|c|c|cccccc|}
\hline
\textbf{Model}                      & \textbf{Emphasis Method} & \multicolumn{6}{c|}{\textbf{Natural Qustions}}                                                                                                                                   \\ \hline
\textbf{}                           & \textbf{}                & \multicolumn{3}{c|}{\begin{tabular}[c]{@{}c@{}}Question\\ First\end{tabular}}                     & \multicolumn{3}{c|}{\begin{tabular}[c]{@{}c@{}}Context\\ First\end{tabular}} \\ \hline
                                    &                          & \multicolumn{1}{c|}{No Emphasis} & \multicolumn{1}{c|}{Q}    & \multicolumn{1}{c|}{C}             & \multicolumn{1}{c|}{No Emphasis} & \multicolumn{1}{c|}{Q}    & C             \\ \hline
\multirow{5}{*}{Falcon-7B}          &                          & \multicolumn{1}{c|}{17.0}        & \multicolumn{1}{c|}{}     & \multicolumn{1}{c|}{}              & \multicolumn{1}{c|}{40.2}        & \multicolumn{1}{c|}{}     &               \\ \cline{2-8} 
                                    & *                        & \multicolumn{1}{c|}{}            & \multicolumn{1}{c|}{10.2} & \multicolumn{1}{c|}{12.8}          & \multicolumn{1}{c|}{}            & \multicolumn{1}{c|}{25.0} & 38.6          \\ \cline{2-8} 
                                    & "                        & \multicolumn{1}{c|}{}            & \multicolumn{1}{c|}{17.0} & \multicolumn{1}{c|}{36.8}          & \multicolumn{1}{c|}{}            & \multicolumn{1}{c|}{38.0} & \textbf{42.4} \\ \cline{2-8} 
                                    & mark                     & \multicolumn{1}{c|}{}            & \multicolumn{1}{c|}{11.0} & \multicolumn{1}{c|}{34.0}          & \multicolumn{1}{c|}{}            & \multicolumn{1}{c|}{34.4} & 41.4          \\ \cline{2-8} 
                                    & emphasis                 & \multicolumn{1}{c|}{}            & \multicolumn{1}{c|}{9.8}  & \multicolumn{1}{c|}{30.8}          & \multicolumn{1}{c|}{}            & \multicolumn{1}{c|}{36.0} & 40.2          \\ \hline
\multirow{5}{*}{Falcon-7B Instruct} &                          & \multicolumn{1}{c|}{24.4}        & \multicolumn{1}{c|}{}     & \multicolumn{1}{c|}{}              & \multicolumn{1}{c|}{39.8}        & \multicolumn{1}{c|}{}     &               \\ \cline{2-8} 
                                    & *                        & \multicolumn{1}{c|}{}            & \multicolumn{1}{c|}{24.6} & \multicolumn{1}{c|}{17.4}          & \multicolumn{1}{c|}{}            & \multicolumn{1}{c|}{20.8} & 40.6          \\ \cline{2-8} 
                                    & "                        & \multicolumn{1}{c|}{}            & \multicolumn{1}{c|}{29.0} & \multicolumn{1}{c|}{34.2}          & \multicolumn{1}{c|}{}            & \multicolumn{1}{c|}{16.6} & 36.2          \\ \cline{2-8} 
                                    & mark                     & \multicolumn{1}{c|}{}            & \multicolumn{1}{c|}{25.0} & \multicolumn{1}{c|}{\textbf{47.2}} & \multicolumn{1}{c|}{}            & \multicolumn{1}{c|}{16.0} & 39.8          \\ \cline{2-8} 
                                    & emphasis                 & \multicolumn{1}{c|}{}            & \multicolumn{1}{c|}{16.2} & \multicolumn{1}{c|}{42.6}          & \multicolumn{1}{c|}{}            & \multicolumn{1}{c|}{12.6} & 38.6          \\ \hline
\multirow{5}{*}{MPT-7B}             &                          & \multicolumn{1}{c|}{17.0}        & \multicolumn{1}{c|}{}     & \multicolumn{1}{c|}{}              & \multicolumn{1}{c|}{43.5}        & \multicolumn{1}{c|}{}     &               \\ \cline{2-8} 
                                    & *                        & \multicolumn{1}{c|}{}            & \multicolumn{1}{c|}{20.5} & \multicolumn{1}{c|}{18.5}          & \multicolumn{1}{c|}{}            & \multicolumn{1}{c|}{49.0} & \textbf{53.5} \\ \cline{2-8} 
                                    & "                        & \multicolumn{1}{c|}{}            & \multicolumn{1}{c|}{16.0} & \multicolumn{1}{c|}{42.0}          & \multicolumn{1}{c|}{}            & \multicolumn{1}{c|}{36.0} & 37.5          \\ \cline{2-8} 
                                    & mark                     & \multicolumn{1}{c|}{}            & \multicolumn{1}{c|}{34.5} & \multicolumn{1}{c|}{49.5}          & \multicolumn{1}{c|}{}            & \multicolumn{1}{c|}{29.0} & 46.0          \\ \cline{2-8} 
                                    & emphasis                 & \multicolumn{1}{c|}{}            & \multicolumn{1}{c|}{17.5} & \multicolumn{1}{c|}{37.5}          & \multicolumn{1}{c|}{}            & \multicolumn{1}{c|}{38.0} & 52.0          \\ \hline
\multirow{5}{*}{MPT-7B Instruct}    &                          & \multicolumn{1}{c|}{25.0}        & \multicolumn{1}{c|}{}     & \multicolumn{1}{c|}{}              & \multicolumn{1}{c|}{13.0}        & \multicolumn{1}{c|}{}     &               \\ \cline{2-8} 
                                    & *                        & \multicolumn{1}{c|}{}            & \multicolumn{1}{c|}{26.7} & \multicolumn{1}{c|}{20.2}          & \multicolumn{1}{c|}{}            & \multicolumn{1}{c|}{32.0} & 14.0          \\ \cline{2-8} 
                                    & "                        & \multicolumn{1}{c|}{}            & \multicolumn{1}{c|}{15.7} & \multicolumn{1}{c|}{29.2}          & \multicolumn{1}{c|}{}            & \multicolumn{1}{c|}{8.25} & 12.5          \\ \cline{2-8} 
                                    & mark                     & \multicolumn{1}{c|}{}            & \multicolumn{1}{c|}{15.0} & \multicolumn{1}{c|}{26.2}          & \multicolumn{1}{c|}{}            & \multicolumn{1}{c|}{15.0} & 13.0          \\ \cline{2-8} 
                                    & emphasis                 & \multicolumn{1}{c|}{}            & \multicolumn{1}{c|}{20.5} & \multicolumn{1}{c|}{\textbf{40.7}} & \multicolumn{1}{c|}{}            & \multicolumn{1}{c|}{20.5} & 13.0          \\ \hline
\multirow{5}{*}{Llama-13B}          &                          & \multicolumn{1}{c|}{28.4}        & \multicolumn{1}{c|}{}     & \multicolumn{1}{c|}{}              & \multicolumn{1}{c|}{58.6}        & \multicolumn{1}{c|}{}     &               \\ \cline{2-8} 
                                    & *                        & \multicolumn{1}{c|}{}            & \multicolumn{1}{c|}{27.4} & \multicolumn{1}{c|}{27.2}          & \multicolumn{1}{c|}{}            & \multicolumn{1}{c|}{41.2} & 55.8          \\ \cline{2-8} 
                                    & "                        & \multicolumn{1}{c|}{}            & \multicolumn{1}{c|}{30.4} & \multicolumn{1}{c|}{55.0}          & \multicolumn{1}{c|}{}            & \multicolumn{1}{c|}{52.0} & 57.0          \\ \cline{2-8} 
                                    & mark                     & \multicolumn{1}{c|}{}            & \multicolumn{1}{c|}{23.4} & \multicolumn{1}{c|}{36.8}          & \multicolumn{1}{c|}{}            & \multicolumn{1}{c|}{41.6} & 60.0          \\ \cline{2-8} 
                                    & emphasis                 & \multicolumn{1}{c|}{}            & \multicolumn{1}{c|}{26.4} & \multicolumn{1}{c|}{53.4}          & \multicolumn{1}{c|}{}            & \multicolumn{1}{c|}{49.4} & \textbf{60.8} \\ \hline
\end{tabular}
\caption{Analysis of newer models, two of which are instruction-tuned, where all models are evaluated using their maximum context length (up to 4k).}
\label{additional_models}
\end{table}

In addition to our main results, we also add an analysis of five more LLMs, all of which were published in 2023 or afterwards and contain between 7B and 13B parameters. Two of the five additional LLMs were instruction-tuned, to evaluate whether such tuning affect the performance change due to different emphasis methods. Lastly, all five of the additional models were evaluated using their maximum context size (up to 4k). Our results can be seen in Table \ref{additional_models}.

Notably, 1) Our results still hold: A) the ordering of inputs plays a crucial role in all models’ performances, where putting the context first strongly improves performance; B) emphasis methods also improve models’ performances. 2) The context size does not play a role in the results, in the sense that our initial results and conclusions still hold. 3) Instruction-tuned models are also susceptible to input order and emphasis methods.

\section{Conclusion}
% In the last decade, there has been notable advancement in natural language processing. However, some practices have become established without thorough evaluation due to the rapid pace of development. 
Focusing on reading comprehension, we evaluate 1) how the order of the question and context affects model performance; and 2) whether emphasizing either the question, the context, or both enhances performance. 
% Here, we investigate two cases related to reading comprehension: the impact of input order -- question and context -- and the effectiveness of emphasizing either the question, context, or both. 
Experimenting with 9 LLMs across multiple datasets, we find that presenting the context before the question improves model performance, with an accuracy increase of up to $31\%$. Furthermore, emphasizing the context yields superior results compared to emphasizing the question, and in general, emphasizing parts of the input is particularly effective for addressing questions that models lack the parametric knowledge to answer. 
% and is particularly effective for addressing questions that models lack the parametric knowledge to answer. 
% We experiment with both prompt-based and attention-based emphasis approaches and find that the best method is surprisingly simple, as it only requires an additional few tokens to be concatenated to the input and results in an accuracy improvement of up to $36\%$, enabling models to outperform other models that are ten times larger.

\section*{Limitations}
While we try to be comprehensive in our comparisons, we only evaluate one approach to represent the question -- “Question: <q>”, and context: “Context: <c>”. However, as discussed in the Section \ref{related_work}, many other approaches exist. That being said, our goal is not to find the best method, but to highlight the issue that exists in the first place, which is the lack of standardization. Additionally, while we focus on reading comprehension, it is an open question if the emphasis methods and ordering also affect other domains or much larger LLMs (e.g., 70B+ parameters).

\section*{Ethics Statement}
The motivation for this paper is to highlight the issue that exists in the lack of standardization of input presentation in reading comprehension, and to show that emphasizing parts of the inputs can be beneficial. We believe that it is crucial that future work continues to evaluate and improve models' performance using different settings so they can be safely used in practical scenarios. 

\section*{Acknowledgments}
We thank the reviewers for their comments and great suggestions. The authors acknowledge financial support from NIH grants OT2TR003422 and R01LM013400.

% \appendix
% \section{Appendix}
% \subsection{Additional Models with Maximum Context Size}

% \label{max_context_size}

\bibliography{anthology,custom}
\bibliographystyle{acl_natbib}

\end{document}